%% file: revised_aistats_paper.tex
\documentclass[twoside]{article}

\usepackage[accepted]{aistats2025}
\usepackage{xr}
\usepackage[utf8]{inputenc} 
\usepackage[T1]{fontenc}    
\usepackage{hyperref} 
\usepackage{ifthen}
\usepackage{url}            
\usepackage{booktabs}       
\usepackage{amsfonts}       
\usepackage{nicefrac}       
\usepackage{microtype}      
\usepackage{xcolor}         
\usepackage[ruled,linesnumbered]{algorithm2e}
\usepackage{booktabs}
\usepackage{enumerate}
\usepackage{subfigure}
\usepackage{graphicx}[realboxes]
\usepackage{multirow}
\usepackage{enumitem}

\usepackage{dsfont}
\input{preamble}

\newcommand{\supp}{\mathrm{supp}}

\newcommand{\KL}{\mathrm{KL}}
\newcommand{\kl}{\mathrm{kl}}
\newcommand{\Alt}{\mathrm{Alt}}

\ifodd 1
\newcommand{\cssc}[1]{{\color{blue}(Chengshuai: #1)}}

\else
\newcommand{\cssc}[1]{}
\fi

\ifodd 1
\newcommand{\cssr}[1]{{\color{blue}#1}}
\else
\newcommand{\cssr}[1]{#}
\fi

\ifodd 1

\else

\fi

\ifodd 1
\newcommand{\congc}[1]{{\color{red}(Cong: #1)}}
\else
\newcommand{\congc}[1]{}
\fi
\usepackage[round]{natbib}

\allowdisplaybreaks

\begin{document}

%

%

\twocolumn[

\aistatstitle{Cost-Aware Optimal Pairwise Pure Exploration}

\aistatsauthor{ Di Wu \And Chengshuai Shi \And Ruida Zhou \And  Cong Shen }

\aistatsaddress{ University of Virginia \And University of Virginia \And  University of California,\\ Los Angeles \And University of Virginia } 
]

\begin{abstract}
  Pure exploration is one of the fundamental problems in multi-armed bandits (MAB). However, existing works mostly focus on specific pure exploration tasks, without a holistic view of the general pure exploration problem. This work fills this gap by introducing a versatile framework to study pure exploration, with a focus on identifying the pairwise relationships between targeted arm pairs. Moreover, unlike existing works that only optimize the stopping time (i.e., sample complexity), this work considers that arms are associated with potentially different costs and targets at optimizing the cumulative cost that occurred during learning. Under the general framework of pairwise pure exploration with arm-specific costs, a performance lower bound is derived. Then, a novel algorithm, termed CAET (\textbf{C}ost-\textbf{A}ware Pairwise \textbf{E}xploration \textbf{T}ask), is proposed. CAET builds on the track-and-stop principle with a novel design to handle the arm-specific costs, which can potentially be zero and thus represent a very challenging case. Theoretical analyses prove that the performance of CAET approaches the lower bound asymptotically. Special cases are further discussed, including an extension to regret minimization, which is another major focus of MAB. The effectiveness and efficiency of CAET are also verified through experimental results under various settings.
\end{abstract}

\section{INTRODUCTION}\label{sec:intro}

The study of multi-armed bandits (MAB), which captures the essence of sequential decision-making processes, has a long and rich history \citep{thompson1933likelihood, lattimore2020bandit}. Despite being a simple model, MAB has found wide-ranged applications including online recommendations \citep{li2010contextual}, wireless communications \citep{gai2010learning}, resource allocation \citep{liu2021pond}, clinical trails \citep{aziz2021multi}, etc. The targets of learning in MAB can be generally categorized as regret minimization \citep{auer2002finite} and pure exploration \citep{audibert2010best}. 

The most classical pure exploration setting is best arm identification (BAI) \citep{garivier2016optimal,kaufmann2016complexity}, whose extensions have also been broadly studied, e.g., top arms identification \citep{kalyanakrishnan2012pac, zhou2022approximate}. However, taking a closer look at these studies on pure exploration, we recognize that there are two major limitations. First, they are all confined to one specific kind of pure exploration task, while a general framework is still lacking. As a result, whenever a different task is of interest (e.g., identifying the ranking of the arms according to their expected rewards), a new algorithm needs to be developed. Second, they commonly treat all arms equally by setting the target as minimizing the total times of arm pulling (i.e., sample complexity). However, in real-world applications, the costs of sampling different arms may differ. For example, in clinical trials (which is further discussed in Sec.~\ref{subsec:motivation}), the prices of varying candidate vaccines or treatments are mostly different from each other.

This work is motivated by these limitations and targets to provide a more comprehensive study of the pure exploration problem. The detailed contributions are summarized as follows:

\begin{itemize}[noitemsep,topsep=0pt,leftmargin = *]
\item A general framework is established to study pure exploration tasks, with a focus on identifying the pairwise relationship between arms. Furthermore, arm-specific costs are introduced so that the ultimate target is to minimize the cumulative costs during learning. This framework is versatile, subsuming many representative pure exploration tasks, e.g., best arm identification and ranking identification.
\item Through the lens of this framework, unified studies on pure exploration are performed. Importantly, a generic lower bound is derived to characterize the fundamental learning limit, capturing the optimal proportions of costs to be assigned to each arm.
\item Under the pairwise pure exploration framework, a novel $\delta$-PAC algorithm, termed CAET (\textbf{C}ost-\textbf{A}ware Pairwise \textbf{E}xploration \textbf{T}ask), is proposed. It builds on the philosophy of track-and-stop from the study of BAI \citep{garivier2016optimal} with specifically crafted designs. In particular, a novel \textit{forced exploration} method is introduced so that the sampling proportion between arms with zero and non-zero costs can be carefully balanced. We note that none of the previous works have tackled a pure exploration problem with zero-cost arms, to the best of our knowledge.
\item Theoretical analyses demonstrate that CAET is capable of approaching the lower bound in the asymptotic regime, illustrating its optimality. In particular, to obtain the desired performance guarantee, a strong result on the convergence of the empirical sample proportion to the optimal proportion is established.
\item We further extend the discussion to the objective of regret minimization, which under the explore-then-commit scheme can be viewed as a BAI problem with the arms' sub-optimality gaps as costs. We demonstrate that in this case, CAET can still achieve asymptotically optimal performance.
\end{itemize}

\section{RELATED WORKS}\label{sec:relatedwork}
In the long history of MAB research, different kinds of pure exploration tasks have been studied, where the problem of \textbf{best arm identification (BAI)} is arguably the most basic one. As a result, most of the existing studies focus on BAI, e.g., \citet{even2006action,gabillon2012best,kaufmann2016complexity,garivier2016optimal,shen2019universal}. A few extensions have also been investigated. For example, \citet{kalyanakrishnan2012pac,kalyanakrishnan2010efficient, chen2017nearly, zhou2022approximate} have studied the identification of the \textit{top few arms}, instead of only the best one. Also, \citet{gabillon2011multi} has initiated the investigation on identifying the optimal arm in each pre-specified arm group (which is often referred to as the multi-bandit BAI). However, as mentioned in Section~\ref{sec:intro}, all of these works are confined to one specifically formulated pure exploration problem, while our work provides a general framework that can subsume the previous settings. 
In addition, most existing works, including the above-mentioned ones, target at optimizing the sample complexity, where all arm pulls are treated equally. This work, however, considers that the arms are associated with potentially different costs with a target of optimizing the cumulative cost. To the best of our knowledge, this is the first time that a general pure exploration framework with arm-specific cost has been established. 

In the following, we discuss a few key related works. \citet{chen2017nearlyoptimalsamplingalgorithms} studied the so-called ``general sampling problem'' in the setting of pure exploration, and provided a lower and upper bound. However, it is still focusing on the uniform cost and many applications with arm-specific cost will be limited, in the meantime, our work generalizes the bandit's framework to pure exploration with the arm-specific cost and provides the optimal lower bound and a near-optimal algorithm. It is noted that the work of \citet{kanarios2024cost} touches upon \textbf{arm-specific costs} in BAI. In contrast, our work not only studies a more general framework but also considers more general non-negative costs. Furthermore, the investigation of \citet{kanarios2024cost} is restricted to strictly positive costs, which limits its applicability, e.g., it cannot handle regret minimization as our extension in Section~\ref{sec:regretmin}. We particularly note that the \textit{zero-cost arms} introduce significant challenges for both design and analysis, which will be illustrated in later discussions. Also, the recent work \citet{qin2025cost} involves costs in the study of BAI, while focusing on finding the arm with the highest reward-to-cost ratio.
Moreover, another recent work of \citet{zhang2023fast} studies how to leverage the explore-then-commit scheme to perform regret minimization, which is similar to the extension in Section~\ref{sec:regretmin}. The differences between \citet{zhang2023fast} and this work are further discussed in Section~\ref{sec:regretmin}.

Here we also discuss the setting of \textbf{dueling bandits} \citep{du2022duelingbanditstwoduelingmultidueling,yue2012k,dudik2015contextual}, highlighting its differences from the pairwise pure exploration task in this work. In particular, for \textbf{dueling bandits}, a pairwise \emph{feedback} is obtained by selecting two arms at each time step. The pairwise pure exploration task studied in our work is still under the canonical bandit setting, where an arm is pulled at each time step. The reason we name it a pairwise exploration task is that many pure exploration tasks can be exactly represented via pairwise comparisons. For example, in the BAI problem, to guarantee that arm $a^*$ is the best arm, it suffices to show that arm $a^*$ has a larger reward in all $K-1$ pairwise comparisons between arms $(a^*, a)$ for each $a \not= a^*$. To this end, to tackle the pure exploration tasks, it is sufficient to conduct these elementary pairwise comparisons. The class of tasks studied in this paper -- \emph{pairwise pure exploration task} -- is rigorously defined in Definition~\ref{def:pairwise}.

\section{PROBLEM FORMULATION}\label{sec:problem}
\subsection{Bandits with Arm-Dependent Costs}

In this work, we consider a multi-armed bandits (MAB) model with a set of $K$ arms, denoted as $\calA=\{1,2,\ldots,K\}$. As in the canonical MAB studies, each arm $a\in \calA$ is associated with a reward distribution $\pi_a$ whose expectation is denoted as $\mu_a$. Moreover, each arm $a\in \calA$ is associated with a cost distribution $\nu_a$, whose expectation is denoted as $c_a$. To reflect the nature of costs, we assume that $c_a \geq 0$ for all $a\in \calA$. The introduction of the cost distribution allows us to define a broad class of cost-aware pairwise pure exploration tasks later. To ease the exposition, we denote $\boldsymbol{\mu}:= (\mu_1, \mu_2, \ldots, \mu_K)$ and $\boldsymbol{c}:= (c_1, c_2, \ldots, c_K)$.

In this model, at each time $t$, an agent plays an arm $A_t\in \mathcal{A}$. Then, she simultaneously receives a reward $X_t$ and cost value $C_t$ independently sampled from the probability distribution $\pi_{A_t}$ and $\nu_{A_t}$ respectively.

In the following discussion, we focus on the standard scenario \citep{garivier2016optimal} with the reward distributions belonging to a canonical exponential family, defined as 
\begin{align*}
    \mathcal{P}=\left\lbrace (\pi_\theta)_{\theta\in\Theta}:\frac{d\pi_\theta}{d\xi}=\exp(\theta x-b(\theta)) \right\rbrace,
\end{align*}
where $\Theta\in \Rb$, $\xi$ is some reference measure on $\Rb$, and $b:\Theta\rightarrow\Rb$ is a convex, twice differentiable function. A distribution $\pi_\theta\in \mathcal{P}$ can be parameterized by its expectation $\dot{b}(\theta)$ and for every $\mu\in \dot{b}(\Theta)$, we denote by $\pi^\mu$ the unique distribution in $\mathcal{P}$ with expectation $\mu$. Then, the bandit model with reward distributions $\boldsymbol{\pi}=(\pi_1,\ldots,\pi_K)$ can be represented by its means $\bmu=(\mu_1,\ldots,\mu_K)$.

\subsection{The Pairwise Pure Exploration Problem}\label{sec:definition}

Denote by $S_K$ the permutation group of $\calA$, a pure exploration task $(\calG, \varphi)$ can be defined through a partition $\calG=\{\mathcal{G}_1, \ldots, \mathcal{G}_M \}$ of $S_{K}$ (i.e., $\mathcal{G}_{i} \cap\ \mathcal{G}_{j} = \emptyset$ and $\bigcup_{m = 1}^M \mathcal{G}_m = S_K$) and a mapping $\varphi:\Rb^K\rightarrow [M]$ that outputs an index within the partition $\varphi(\boldsymbol{\mu})= m \in [M]$ for any expected reward vector $\boldsymbol{\mu} \in \mathbb{R}^K$. 

We are interested in a sufficiently expressive and representative subclass of all the pure exploration tasks, referred to as the \textbf{pairwise exploration tasks}. To ease the notation, denote by $\sigma^{\bmu}$ the index of descending order of $\bmu$, i.e., $\bmu_{\sigma^{\bmu}(1)} > \bmu_{\sigma^{\bmu}(2)} > \cdots > \bmu_{\sigma^{\bmu}(K)}$.\footnote{We mainly focus on the $\bmu$ without ties, and discuss the case with ties for different tasks individually.} Then, $(\sigma^{\bmu})^{-1}(j) = k$ indicates that arm-$j$ of $\bmu$ is the $k$-th largest arm, and $(\sigma^{\bmu})^{-1}(j) < (\sigma^{\bmu})^{-1}(i)$ indicates the reward of arm-$j$ is larger than that of arm-$i$. Let $\mathcal{B}_{ij} := \{\sigma \in S_K: \sigma^{-1}(i) < \sigma^{-1}(j)\}$. The pairwise exploration task can be defined as follows:
\begin{definition}[Pairwise Exploration Task]\label{def:pairwise}
A task $(\calG, \varphi)$ is a pairwise exploration task if for any $\calG_m \in \calG$, there exists a subset $\mathcal{I}_m \subseteq \{(i, j):1 \leq i \not= j \leq K\}$ such that $\calG_m = \cap_{(i, j) \in \mathcal{I}_m} \mathcal{B}_{ij}$ and $\varphi(\bmu) = m$ if $\sigma^{\bmu} \in \mathcal{G}_m$. 
\end{definition}

Intuitively, we can interpret the pairwise exploration task as the task that can be solved via confirming a set of binary comparisons among $\bmu$. The intersection $\cap_{(i,j) \in \mathcal{I}} \mathcal{B}_{ij}$ indicates checking the condition of $\land_{(i,j) \in \mathcal{I}} ( \sigma^{\bmu}(i) < \sigma^{\bmu}(j) ) = \land_{(i,j) \in \mathcal{I}}\{\text{arm-$i$ is better than arm-$j$ in }\bmu\}$. 

The class of pairwise exploration tasks are sufficiently large that contains many tasks of interest as special cases. We give some examples below.

\begin{example}[Ranking Identification]
    Considering the pure exploration problem $(\mathcal{G}, \varphi)$ of identifying the rank of all the arms with respect to their expected rewards. Thus $M = K!$ and each $\mathcal{G}_m \in \mathcal{G}$ is a singleton containing some $\sigma$. It is a pairwise exploration task since $\{\sigma\} = \cap_{i = 1}^{K-1} \mathcal{B}_{\sigma^{-1}(i)\sigma^{-1}(i+1)}$.
\end{example}
It can be observed that ranking identification is a particularly challenging task because $M = K!$, even though the minimum number of binary comparisons needed to identify a ranking is only $K-1$. 
\begin{example}[BAI]
    In the pure exploration problem $(\mathcal{G}, \varphi)$ of identifying the best arm, $M = K$ and $\mathcal{G}_k = \cap_{j \not= k} \mathcal{B}_{kj}$, which is a pairwise exploration task. 
\end{example}
\begin{example}[Best-$m$-arms Identification]
    Considering the pairwise pure exploration problem $(\mathcal{G}, \varphi)$ of identifying the set of $m$ arms with the highest rewards. We can take $M = {K\choose m}$ and for each $\boldsymbol{i}=(i_1,\ldots,i_m)$ and $\mathcal{G}_{\boldsymbol{i}} = \cap_{k\notin\{i_1,\ldots,i_m\}} \cap_{t=1}^m \mathcal{B}_{i_tk}$.
\end{example}
Since for pairwise exploration tasks $(\mathcal{G}, \varphi)$, $\varphi$ is defined explicitly by $\mathcal{G}$, we omit $\varphi$ and use $\mathcal{G}$ to represent the exploration tasks for simplicity.

\subsection{Cost-Aware Pure Exploration with Fixed Confidence}\label{sec:notation}
Denoting $\mathcal{F}_t=\sigma(X_1,C_1,\ldots,X_t,C_t)$ as the $\sigma$-field generated by the observations (i.e., both rewards and costs) up to time $t$, a strategy should generally have the following three components: 
\begin{itemize}[noitemsep,topsep=0pt,leftmargin = *]
    \item a sampling rule $(A_t)_t$, where $A_t$ is $\calF_t$ measurable;
    \item a stopping rule $\tau$, where $\tau$ is a stopping time with respected to $\calF_t$; and
    \item an $\calF_\tau$-measurable decision rule $\hat{m}_\tau$. 
\end{itemize}
This work focuses on the so-called \textit{fixed-confidence} setting and targets at minimizing the cumulative costs. Specifically, given a confidence parameter $\delta\in [0,1]$, the designed strategy should guarantee that the decision rule outputs the correct index $m$ with probability at least $1-\delta$, i.e.,
\begin{align*}
    \mathbb{P}(\hat{m}_\tau \neq m) \leq \delta,
\end{align*}
while minimizing the expected incurring cost defined as
\begin{align}\label{eqn:cost_def}
    \mathbb{E}[f(\boldsymbol{c}, \boldsymbol{\mu};\tau)]=\sum_{a\in \calA} c_a \mathbb{E}[N_a(\tau)],
\end{align}
where $N_a(\tau) = \sum_{t\leq \tau}\mathds{1}\{A_t = a\}$. As the stopping rule $\tau$ in general would be related to the required confidence $\delta$, we adopt the notation $\tau_\delta$ to specify this dependency.

It can be realized that the canonical study of pure exploration problems targets at ensuring the identification correctness while minimizing the sample complexity $\tau$, which can be interpreted as a specific case of an all-one cost vector $\bc=(1,\ldots,1)$ in Eqn.~\eqref{eqn:cost_def}. The consideration of the general pure exploration tasks and the flexible cost vector $\bc$ largely broaden the scope of the previous pure exploration studies. In particular, there may exist $c_i = 0$ in the cost vector $\bc$, which significantly increases the technical difficulty in the strategy design and analysis, as will be evident later.

\paragraph{Preliminaries and Notations}
For two probability distributions $p$ and $q$, we denote their Kullback-Leibler (KL) divergence as $\KL(p,q)$. As has been stated in \cite{Capp__2013}, the KL divergence from two distributions $\pi_\theta$ and $\pi_{\theta'}$  in the exponential family, with expectations $\mu$ and $\mu'$, induces a divergence function $d$ on $\dot{b}(\Theta)$ defined as $$d(\mu,\mu')=\KL(\pi_\theta,\pi_{\theta'})=b(\theta')-b(\theta)-\dot{b}(\theta)(\theta'-\theta).$$
Specifically, when two reward distributions are Gaussian with expectations $x,y$ and variance $\sigma^2$, it holds that $d(x,y)=(x-y)^2/(2\sigma^2)$; when two reward distributions are Bernoulli with expectations $x,y$, it holds that $d(x,y)=\kl(x,y)=x\log(x/y)+(1-x)\log ((1-x)/(1-y))$.

We denote $P(\boldsymbol{c}): =\{a \in \calA: c_a>0\}$ as the set of arms associated with non-zero expected costs and $N(\boldsymbol{c}):=\{a \in \calA: c_a=0\}$ as the set of arms associated with zero expected costs. Also, for any set $\calI$ defined in Definition~\ref{def:pairwise}, we define its support set $\supp(\calI)=\{i: i\in \calA, \text{$\exists j\in \calA/\{i\}$, s.t. $(i,j)\in \calI$}\}$. For exploration task $(\calG,\varphi), \calG=\{\calG_1,\ldots,\calG_M\}$ and we denote $\calI_m$ be the index set of intersection: $\calG_m=\cap_{(i,j)\in\calI_m}\calB_{ij}$. For bandit model $\bmu$, with $\varphi(\bmu)=m$, we further define $P_{\calG}(\bc,\bmu)=P(\bc)\cap\supp(\calI_m)$, $N_{\calG}(\bc,\bmu)=N(\bc)\cap\supp(\calI_m)$, and $\calI(\bmu)=\calI_m=\calI_{\varphi(\bmu)}$. Furthermore, a notation table is provided in Appendix~\ref{app:notation} to facilitate reading. 

\subsection{Motivation Applications}\label{subsec:motivation}
In the following, two motivation applications are discussed to highlight the practical relevance of the cost-aware pure exploration problem formulated above.

First, in clinical trials (e.g., identifying the most effective vaccine or treatment among multiple candidates), the costs of different vaccines or treatments can vary significantly. Therefore, adopting a cost-aware design, as proposed in this work, is advantageous, as it can lead to substantial budget savings compared to previous approaches that assume uniform costs. In addition, self-healing (using placebo or no active intervention) often serves as a crucial baseline for evaluating the efficacy of medical treatments. Such baselines inherently have no additional cost and can be modeled effectively as zero-cost arms as considered in this work.

Similarly, in wireless communications, different channels allow for varying communication rates, while also requiring different transmit power levels to communicate. In the task of finding the best channel with the highest communication rate, this work can further minimize the overall power (i.e., cumulative costs) consumed during the process, instead of simply measuring the overall channel usage as in previous works.

\section{LOWER BOUND}\label{sec:sample}

In this section, we establish the fundamental limits of the general pure exploration problems. Let $(\calG,\varphi)$ be the pairwise exploration task that is concerned. We first define $\mathcal{S}$ as the set of bandit models whose expected rewards $\boldsymbol{\mu}$ satisfying that $\varphi(\bmu)$ belongs to exactly one partition $\calG_m$. We say a strategy is $\delta$-PAC if, for every $\boldsymbol{\mu}\in \mathcal{S}$, it satisfies that $\Pb_{\boldsymbol{\mu}}(\tau_\delta<\infty)=1$ and $\Pb_{\boldsymbol{\mu}}(\hat{m}_{\tau_\delta} \neq m)\leq \delta$. Furthermore, we introduce the notion of an \emph{alternative set} as 
\begin{align}\label{for:altalt}
    \Alt(\boldsymbol{c},\boldsymbol{\mu})=\{&\boldsymbol{\lambda} \in S: \varphi(\boldsymbol{\lambda})\neq \varphi(\bmu),\notag\\
    &\lambda_i=\mu_i, \forall i \notin P_\calG(\bc,\bmu)\}, 
\end{align}

and denote the following distribution set
\begin{align}\label{for:distributiondis}
    \Sigma_{P_{\calG}(\bc,\bmu)}:=\{&\boldsymbol{\omega}\in \Delta_{K}: \omega_i=0, \omega_j>0, \notag\\
    &\forall i\notin P_{\calG}(\bc,\bmu), \forall j\in P_{\calG}(\bc,\bmu)\},
\end{align}
where $\Delta_{K}$ denotes the probability simplex of $K-1$ dimensions. Then, the following lower bound can be established.
\begin{theorem}\label{the:lower-bound}
    Let $\delta \in (0,1)$ and give a pairwise exploration task $(\calG,\varphi)$. For any $\delta$-PAC strategy and any bandit model $\boldsymbol{\mu}\in S$, the following holds:
    \begin{align*}
        \mathbb{E}[f(\boldsymbol{c}, \boldsymbol{\mu};\tau_\delta)]\geq T^*(\boldsymbol{c},\boldsymbol{\mu})\kl(\delta,1-\delta),
    \end{align*}
    where
    \begin{align}\label{formula:T}
        \begin{split}
            &T^*(\boldsymbol{c},\boldsymbol{\mu})^{-1}=\\
            &\sup\limits_{\omega \in \Sigma_{P_{\calG}(\bc,\bmu)}}\inf\limits_{\boldsymbol{\lambda}\in \Alt(\boldsymbol{c},\boldsymbol{\mu})}\left\{\sum\limits_{a \in P_{\calG}(\bc,\bmu)} \omega_a \frac{d(\mu_a,\lambda_a)}{c_a}\right\}.
        \end{split}
    \end{align}
\end{theorem}
    As $\kl(\delta,1-\delta)\sim\log(1/\delta)$ when $\delta$ goes to zero, this theorem yields the following asymptotic lower bound:
    \begin{align*}
        \liminf\limits_{\delta\rightarrow 0}\frac{\mathbb{E}[f(\boldsymbol{c}, \boldsymbol{\mu};\tau_\delta)]}{\log(1/\delta)}\geq T^*(\boldsymbol{c},\boldsymbol{\mu}).
    \end{align*}
    A non-asymptotic version can be obtained from the inequality $\kl(\delta,1-\delta)\geq \log(1/(2.4\delta))$ as $\mathbb{E}[f(\boldsymbol{c}, \boldsymbol{\mu};\tau_\delta)]\geq T^*(\boldsymbol{c},\boldsymbol{\mu})\log(1/(2.4\delta))$. 
\begin{Proofsketch}
    To establish the lower bound, we first consider a transportation lemma of \citet{kaufmann2016complexity} which relates to the expected number of draws in two bandit models being different under the given exploration task:
    \begin{align*}
        &\forall \boldsymbol{\lambda}\in \Alt(\boldsymbol{c},\boldsymbol{\mu}),\\
        &\sum\limits_{a\in P_\calG(\boldsymbol{c},\bmu)}d(\mu_a,\lambda_a)\Eb[N_a(\tau_\delta)]\geq \kl(\delta,1-\delta),
    \end{align*}
    Considering  $\mathbb{E}^\calG[f(\boldsymbol{c}, \boldsymbol{\mu};\tau_\delta)]:=\sum_{a\in P_\calG(\boldsymbol{c},\bmu)}c_a\mathbb{E}[N_a(\tau_\delta)]$, we have 
    \begin{align*}
    &\kl(\delta,1-\delta)\leq \inf\limits_{\boldsymbol{\lambda}\in \Alt(\boldsymbol{c},\boldsymbol{\mu})}\mathbb{E}^{\calG}[f(\boldsymbol{c}, \boldsymbol{\mu};\tau_\delta)]\cdot \\
    &\quad\quad\quad\quad\quad\left(\sum\limits_{a\in P_\calG(\boldsymbol{c},\bmu)}\frac{c_a\mathbb{E}[N_a]}{\mathbb{E}^\calG[f(\boldsymbol{c}, \boldsymbol{\mu};\tau_\delta)]}\frac{d(\mu_a,\lambda_a)}{c_a}\right)\\
    &\leq \mathbb{E}^\calG[f(\boldsymbol{c}, \boldsymbol{\mu};\tau_\delta)]\cdot\\
    &\quad\sup\limits_{\omega \in \Sigma_{P_\calG(\boldsymbol{c},\bmu)}}\inf\limits_{\boldsymbol{\lambda}\in \Alt(\boldsymbol{c},\boldsymbol{\mu})}\left(\sum\limits_{a \in P_\calG(\boldsymbol{c},\bmu)} \omega_a \frac{d(\mu_a,\lambda_a)}{c_a}\right).
    \end{align*}
Finally, we have the desired result $\mathbb{E}[f(\boldsymbol{c}, \boldsymbol{\mu};\tau_\delta)]\geq \mathbb{E}^\calG[f(\boldsymbol{c}, \boldsymbol{\mu};\tau_\delta)] \geq T^*(\bc,\bmu)\kl(\delta,1-\delta)$. 
\end{Proofsketch}

It can be observed that the supremum in $T^*(\boldsymbol{c},\boldsymbol{\mu})$ is indeed a maximum because, by Lemma~\ref{lem:optgernal}, $\inf_{\boldsymbol{\lambda}\in \Alt(\boldsymbol{c},\boldsymbol{\mu})}\{\sum_{a \in P_{\calG}(\bc,\bmu)} \omega_a d(\mu_a,\lambda_a)/c_a\}$ is a continuous function, and any continuous function defined in a closed set can reach its maximum value. Thus, we define
\begin{align*}
    &\boldsymbol{\omega}^*(\boldsymbol{c},\boldsymbol{\mu}):=\\
    &\argmax\limits_{\boldsymbol{\omega} \in \Sigma_{P_{\calG}(\bc,\bmu)}}\inf\limits_{\boldsymbol{\lambda}\in \Alt(\boldsymbol{c},\boldsymbol{\mu})}\left\{\sum\limits_{a \in P_{\calG}(\bc,\bmu)} \omega_a \frac{d(\mu_a,\lambda_a)}{c_a}\right\},
\end{align*}
which is directly related to the optimal sampling proportion and $\omega_a$ represents the ratio of arm $a$ among all the arms. However, the vector $\boldsymbol{\omega}^*$ is not the desired sample distribution due to the scaling of the cost vector. Thus, the following transform function $G_{\boldsymbol{c}}:\Sigma_{P_{\calG}(\bc,\bmu)}\rightarrow \Sigma_{P_{\calG}(\bc,\bmu)}$ is defined to get the actual sampling distribution from $\boldsymbol{\omega}^*$: for cost vector $\boldsymbol{c}$,
\begin{align*}
    &G_{\boldsymbol{c}}([a_1,\ldots , a_K])= [b_1,\ldots,b_K], \text{where} \\
    &b_i = \begin{cases}
        \frac{a_i/c_i}{\sum_{s\in P_{\calG}(\bc,\bmu)}a_s/c_s} & \text{if } i\in P_{\calG}(\bc,\bmu)\\
        0 & \text{if } i\notin P_{\calG}(\bc,\bmu).
    \end{cases}
\end{align*}
The complete proof of Theorem~\ref{the:lower-bound}, which is given in Appendix~\ref{app:lowerbound}, shows that $\boldsymbol{u}^*(\boldsymbol{c},\boldsymbol{\mu}):=G_{\boldsymbol{c}}(\boldsymbol{\omega}^*(\boldsymbol{c},\boldsymbol{\mu}))$ is the proportion of pulling arms in $P(\bc)\cap\supp(\calI_m)$ that matches this lower bound. Thus, the pulling proportion of the arm in $P(\bc)$ in the optimal algorithm should be with respect to this pulling proportion. Our sampling rule in Section~\ref{sec:TAS} indeed follows this intuition. Additional results on more explicit forms of $T^*(\bc,\bmu)$ can be found in Appendix~\ref{sec:character}.

\section{CAET ALGORITHM}\label{sec:TAS}
We now describe a novel strategy to tackle the cost-aware pairwise exploration problem $\calI$ called CAET (\textbf{C}ost-\textbf{A}ware Pairwise Pure \textbf{E}xploration \textbf{T}ask). Without loss of generality, we consider the scenarios with $\supp(\calI)=\calA $, as otherwise one can exclude the arms not considered in $\calI$ from  $\calA$ to form a new arm set. We will present a sampling rule in Section~\ref{sec:samplingrule}, a stopping rule in Section~\ref{sec:stoppingrule}, and a decision rule in Section~\ref{sec:decisionrule}, whose combination leads to the proposed CAET algorithm that is compactly presented in Algorithm~\ref{algo:CAET}.  In the following discussion, the set $N(\bc)$ will be assigned in advance, and we generally denote the current sample means of expected rewards $\boldsymbol{\mu}$ at time $t$ as $\hat{\bmu}(t)=(\hat{\mu}_1(t),\ldots,\hat{\mu}_K(t))$ where $\hat{\mu}_a(t)=N_a(t)^{-1}\sum_{s\leq t}X_s\mathbf{1}_{\{A_s=a\}}$ and the current sample means of expected costs $\boldsymbol{c}$  at time $t$ as $\hat{\bc}(t)=(\hat{c}_1(t),\ldots,\hat{c}_K(t))$ where $\hat{c}_a(t)=N_a(t)^{-1}\sum_{s\leq t}C_s\mathbf{1}_{\{A_s=a\}}$ for arm $a\in P(\bc)$.

\subsection{Sampling Rule}\label{sec:samplingrule}

We begin the description of the sampling rule with the simple scenario where the expected costs are all positive, i.e., $c_a > 0$ for all $a\in \calA$, or equivalently, $|N(\bc)|=0$. As the optimal sampling proportion for all arms in this case is $\boldsymbol{u}^*(\bc,\bmu)=G_{\bc}(\boldsymbol{\omega}^*(\bc,\bmu))$, one natural idea is to sample according to the plug-in estimate $\boldsymbol{u}^*(\bc_\delta(t),\hat{\bmu}(t))$. Especially, the C-tracking strategy proposed in \citet{garivier2016optimal} can be adopted. 

With the sampling intuition explained under non-zero expected costs, we move to the challenging case with some arms having zero expected costs, i.e., $|N(\bc)| \neq 0$. It can be observed that the introduction of zero costs drastically complicates the problem, as $\boldsymbol{u}^*(\bc_\delta(t),\hat{\bmu}(t))$ (cf. step 4 in Algorithm~\ref{algo:CAET}) only has non-zero elements for arms in $P(\bc)$ and thus does not describe how arms in $N(\bc)$ should be sampled. To overcome this issue and particularly guarantee a finite stopping time, we introduce a proportion $\alpha \in(0,1)$ to perform uniform sampling over arms with zero costs. The arms with non-zero costs share the other $1-\alpha$ pulling proportion and are still sampled by tracking the estimate of the optimal proportion as described before. In particular, we design the following overall distribution to be tracked (cf. step 5 in Algorithm~\ref{algo:CAET}):\footnote{When there is no cost-zero arm, i.e. $N_{\calG}(\bc_\delta(t),\hat{\bmu}(t))=\emptyset$, the sample distribution can be directly set as $\boldsymbol{u}^*(\bc_\delta(t),\hat{\bmu}(t))$.} 
\begin{align*}
    &\boldsymbol{u}_\alpha (\bc_\delta(t),\hat{\bmu}(t))=\\
    &\underbrace{\lv_\alpha(\bc_\delta(t),\hat{\bmu}(t))}_{\text{arms in $N_{\calG}(\bc_\delta(t),\hat{\bmu}(t))$}}+\underbrace{(1-\alpha)\boldsymbol{u}^*(\bc_\delta(t),\hat{\bmu}(t))}_{\text{arms in $P_\calG(\bc_\delta(t),\hat{\bmu}(t))$}},
\end{align*}
where $\bc_\delta(t)=D_\delta(\hat{\bc}(t))$, $D_\delta([a_1,\ldots,a_K])=[b_1,\ldots,b_K]$ with $b_i=a_i$ if $a_i>\gamma_0\log^{-r'}(1/\delta)$ and $b_i=0$ otherwise, where $\gamma_0$ is an arbitrary positive constant and $r'$ is a constant satisfying $0<r'<1/8$. Also, $\lv_\alpha (\boldsymbol{c}',\bmu'):=(i_1,\ldots,i_K)$ with $i_t=\alpha/|N_{\calG}(\boldsymbol{c}',\bmu')|$ if $t\in N_{\calG}(\boldsymbol{c}',\bmu')$ and $i_t=0$ otherwise. It can be observed that the parameter $\alpha$ balances the sampling proportion between the arms with zero and non-zero costs. 
Intuitively, to guarantee a small cumulative cost when $\delta \to 0$, those arms with zero costs should be pulled sufficiently so that their estimations are precise enough to guide the estimation of $\boldsymbol{u}^*(\bc_\delta(t),\hat{\bmu}(t))$ for the other arms. Thus, intuitively, the choice of $\alpha$ should be adaptive in $\delta$ such that $\lim_{\delta\rightarrow 0}\alpha=1$. In particular,  we take $\alpha=1-\log^{-r}(1/\delta)$ in the proposed CAET design for any chosen $0<r<1/2$. More detailed explanations are provided in Section~\ref{sec:convergence}.

Finally, the sampling rule (cf. step 6 in Algorithm~\ref{algo:CAET}) is designed as
\begin{align}\label{for:sampling}
    A_{t+1}\in \argmax\limits_{1\leq a\leq K}\sum\limits_{s=0}^t u_{\alpha, a}^{\epsilon_s}(\bc_\delta(s),\hat{\bmu}(s))-N_a(t),
\end{align}
where $\boldsymbol{u}_\alpha^{\epsilon}(\bc_\delta(t),\hat{\bmu}(t)) = (u_{\alpha, 1}^{\epsilon}(\bc_\delta(t),\hat{\bmu}(t)), \ldots, u_{\alpha, K}^{\epsilon}(\bc_\delta(t),\hat{\bmu}(t)))$ is the $L^\infty$ projection of $\boldsymbol{u}_\alpha(\bc_\delta(t),\hat{\bmu}(t))$  to $\Sigma_K^{\epsilon} =\{(\omega_1,\ldots,\omega_K)\in [\epsilon,1]^K:\sum_{i=1}^K\omega_i=1\}$. This sampling rule guarantees that besides some forced explorations for information collection introduced by the projection, the arms are pulled roughly according to $\boldsymbol{u}_\alpha(\bc_\delta(t),\hat{\bmu}(t))$. For the choice of $\epsilon_t$, we take $\epsilon_t=(K^2+t)^{-1/2}/2$, which leads to the desired performance. One important property of this sampling rule is that the empirical sample proportion is guaranteed to converge to the optimal proportion, which will be discussed in Section~\ref{sec:convergence}.

\begin{algorithm}[tb]
  \SetAlgoLined
  \textbf{Input:} $K$, confidence $\delta$, $\bmu,\boldsymbol{\nu}$\;
  \While{$t\in \mathbb{N},\ \exists (a,b)\in \calI\ s.t., Z_{a,b}(t) \leq\beta(t,\delta)$}{
    Calculate $\hat{\boldsymbol{c}}(s),\hat{\boldsymbol{\mu}}(s),s\leq t$\;
    Calculate $\boldsymbol{u}^*(\bc_\delta(t),\hat{\bmu}(t))=G_{\bc_\delta(t)}(\boldsymbol{\omega}^*(\bc_\delta(t),\hat{\bmu}(t)))$\;
    Calculate $\boldsymbol{u}_\alpha (\bc_\delta(t),\hat{\bmu}(t))=\lv_\alpha(\bc_\delta(t),\hat{\bmu}(t))+(1-\alpha)\boldsymbol{u}^*(\bc_\delta(t),\hat{\bmu}(t))$\;
    Choose action $A_{t+1}\in \argmax_{1\leq a\leq K}\sum_{s=0}^t u_{\alpha, a}^{\epsilon_s}(\bc_\delta(s),\hat{\bmu}(s))-N_a(t),$\;
    Observe $X_{t+1}$ and $C_{t+1}$ and update data\;
    $t=t+1$\;
  }
  \textbf{Output:} Output the $\calG_m$ determined by the stopping rule \Cref{for:stoppingrule}
  \caption{CAET (\textbf{C}ost-\textbf{A}ware Pairwise Pure \textbf{E}xploration \textbf{T}ask)}
  \label{algo:CAET}
\end{algorithm}

\subsection{Stopping Rule}\label{sec:stoppingrule}
For any pair of arms $a,b\in \mathcal{A}$, we consider the \textit{generalized likelihood ratio} as follows:
\begin{align*}
    Z_{a,b}(t):=\log\frac{\max_{\mu_a'\geq\mu_b'}p_{\mu_a'}(\underline{X}^a_{N_a(t)})p_{\mu_b'}(\underline{X}^b_{N_b(t)})}{\max_{\mu_a'\leq\mu_b'}p_{\mu_a'}(\underline{X}^a_{N_a(t)})p_{\mu_b'}(\underline{X}^b_{N_b(t)})},
\end{align*}
where $\underline{X}^a_{N_a(t)}=(X_s:A_s=a,s\leq t)$ contains the observations of arm $a$ available at time $t$, and $p_{\mu}(Z_1,\ldots,Z_n)$ is the likelihood of $n$ i.i.d. observations from $\omega^{\mu}$ as 
\begin{align*}
    p_{\mu}(Z_1,\ldots,Z_n)=\prod\limits_{k=1}^n \exp(\dot{b}^{-1}(\mu)Z_k-b(\dot{b}^{-1}(\mu))).
\end{align*}
For the exponential bandit models, if $\hat{\mu}_a(t)\geq \hat{\mu}_b(t)$, we have 
\begin{align*}
    Z_{a,b}(t)=N_a(t)d(\hat{\mu}_a(t),\hat{\mu}_{a,b}(t))+N_b(t)d(\hat{\mu}_b(t),\hat{\mu}_{a,b}(t)),
\end{align*}
where $$\hat{\mu}_{a,b}(t)\ := \ \frac{N_a(t)}{N_a(t)+N_b(t)}\hat{\mu}_a(t)+\frac{N_b(t)}{N_a(t)+N_b(t)}\hat{\mu}_b(t)$$
is the empirical mean of arms $a,b$. Note that $Z_{a,b}(t)$ is non-negative if and only if $\hat{\mu}_a(t)\geq \hat{\mu}_b(t)$. Thus, in order to identify to which partition $\calG_m$ $\bmu$ belongs, we use $\calI_m$ to represent the index set $\calI$ of $\calG_m=\cap_{(i,j)\in\calI_m}\calB_{ij}$ and check the value of $Z_{a,b}(t)$. The stopping rule can be specified as 
\begin{align}\label{for:stoppingrule}
    \begin{split}
        &\tau_\delta =\\
        &\inf\{t\in N: \exists\calG_m, \forall (a,b)\in\calI_m, Z_{a,b}(t)>\beta(t,\delta)\},
    \end{split}
\end{align}
where the threshold $\beta(t, \delta)$ needs a careful design to ensure the $\delta$-PAC property. Several provably efficient choices of $\beta(t, \delta)$ are provided in Section~\ref{sec:threshold} with the corresponding analyses.

\subsection{Decision Rule}\label{sec:decisionrule}
When the algorithm stops, it means there exists $\calG_m$ such that $\forall (a,b)\in\calI_m, Z_{a,b}(t)>\beta(t,\delta)$ at a certain time $\tau_\delta$. Our decision rule $\hat{m}_\tau$ will output this $\calG_m$. 

\section{UPPER BOUND}\label{sec:upperbound}
In this section, we first analyze the convergence of the empirical sample proportion to the optimal one. Then, a few provably efficient choices of the threshold in the stopping rule are provided. Finally, the expected asymptotic convergence of CAET is presented, which approaches the lower bound in Theorem~\ref{the:lower-bound}, demonstrating the asymptotic optimality of CAET. 

\subsection{Convergence to the Optimal Proportion}\label{sec:convergence}
First, we discuss the convergence of the empirical sampling proportion to the optimal proportion, where the existence of zero-cost arms introduces analytical challenges. Especially, when $N_\calG(\bc,\bmu)=\emptyset$, the optimal proportion $\bu^*(\bc,\bmu)$ does not change with the choice of $\delta$. Since $\lim_{\delta\rightarrow 0}\tau_\delta=\infty$, Proposition~\ref{pro:time-converge} is sufficient to obtain the convergence guarantee.

However, when there are zero-cost arms in $\supp(\calI(\bmu))$, $N_\calG(\bc,\bmu)\neq\emptyset$, and the optimal proportion $\bu_\alpha(\bc,\bmu)=\lv_\alpha(\bc)+(1-\alpha)\bu^*(\bc,\bmu)$ change with $\delta$ due to the $\delta$-dependency of the choice of $\alpha$. As a result, the key challenge is to prove that
\begin{align*}
    &\Pb\left(\lim\limits_{\delta\rightarrow 0}\frac{N_a(\tau_\delta)}{(1-\alpha)\tau_\delta}=u_a^*\right)=1,\ \  a\in P_\calG(\bc,\bmu); \\
    &\Pb\left(\lim\limits_{\delta\rightarrow 0}\frac{N_a(\tau_\delta)}{\tau_\delta}=\frac{1}{|N(\bc)|}\right)=1,\ \  a\in N_\calG(\bc,\bmu)
\end{align*}
Moreover, for arm $a\in P(\bc)$, the optimal proportion $u_a=(1-\alpha)u^*_a$ will go to zero with $\alpha\to 1$ when $\delta \to 0$, which means the time of $u_a$ under the influence of the $\epsilon^t$-$L^\infty$ projection keeps increasing. Thus, the main difficulty in proving the desired convergence is to handle the balance between the stopping time $\tau_\delta$ and the amount of time that $L^\infty$ has an impact on $u_a$. To tackle the problem, we derive the potential proprieties of a series of random processes and use some adaptive sampling technology to control the whole process, which is novel and of its own merit. The detailed proof is provided in Proposition~\ref{pro:deltaconverge} of Appendix~\ref{appendix:track}.

\subsection{Threshold in the Stopping Rule}\label{sec:threshold}
Parameter $\beta(t,\delta)$ can be viewed as an exploration rate, which requires a careful design to achieve the $\delta$-PAC property. In the following, two provably efficient methods are provided.

\begin{proposition}[Informational Threshold]\label{thm:inf-thre}
    Let $\boldsymbol{\mu}$ be a Bernoulli bandit model. Let $\delta \in (0,1)$. For any sampling strategy, using the stopping rule given in \Cref{for:stoppingrule} on Bernoulli bandits with threshold $\beta(t,\delta)=\log(2tK(K-1)/\delta)$ ensures that for all $\boldsymbol{\mu} \in S$, $\Pb_{\bmu} (\tau_\delta < \infty, \varphi(\hat{\bmu}(\tau_\delta))\neq \varphi(\boldsymbol{\mu}))\leq \delta$. 
\end{proposition}

For the deviational threshold, to guarantee the $\delta$-PAC property in any exponential bandit models, by making use of a deviation result from \cite{Magureanu2014}, we have the following proposition. 

\begin{proposition}[Deviational Threshold]\label{pro:Dev-Thre}
    
    Let $\boldsymbol{\mu}$ be an exponential family of bandit models. Let $\delta \in (0,1)$ and $\theta >1$. There exists a constant $C=C(\theta, K)$ such that for any sampling strategy, using the stopping rule given in \Cref{for:stoppingrule} with threshold $\beta(t,\delta)=\log\left(Ct^\theta/\delta\right)$
    ensures that for all $\bmu \in S$, $\Pb_{\bmu} (\tau_\delta < \infty$, $\varphi(\hat{\bmu}(\tau_\delta))\neq \varphi(\boldsymbol{\mu}))\leq \delta$.
\end{proposition}
The proof of these two propositions can be found in Appendix~\ref{app:section6}.

\subsection{Asymptotic Optimality in Expectation}\label{subsec:opt}
The following theorem establishes the asymptotic optimality of the proposed CAET algorithm.  
\begin{theorem}[Upper Bound]\label{thm:upperbound}
     Let $\theta \in [1,e/2]$ and $r(t)=O(t^\theta)$. Using the stopping rule in \Cref{for:stoppingrule} with an exploration rate $\beta(t,\delta)=\log(r(t)/\delta)$ and the sampling rule in \Cref{for:sampling} with a sample distribution function $\boldsymbol{u}_\alpha$ under $\alpha=1-\log^{-r}(1/\delta)$, we have
    \begin{align*}
        \limsup\limits_{\delta \rightarrow 0}\dfrac{\mathbb{E}[f(\boldsymbol{c}, \boldsymbol{\mu};\tau_\delta)]}{\log(1/\delta)}\leq \theta T^*(\boldsymbol{c},\boldsymbol{\mu}).
    \end{align*}
\end{theorem}
\begin{Proofsketch}
    We provide a brief proof sketch of Theorem~\ref{thm:upperbound}, while the complete proof is given in Appendix~\ref{sec:upperbound}. Intuitively, directly calculating the cumulative cost $\mathbb{E}[f(\boldsymbol{c}, \boldsymbol{\mu};\tau_\delta)]$ is hard, since it needs to handle the arm-specific costs. Thus, by extending \cite{garivier2016optimal}, we can first obtain the following result regarding the sample complexity $\mathbb{E}_{\boldsymbol{\mu}}[\tau_\delta]$:    
    \begin{align}\label{formula:upperbound}
        \limsup\limits_{\delta \rightarrow 0}\dfrac{\mathbb{E}_{\boldsymbol{\mu}}[\tau_\delta]}{\log(1/\delta)}\leq \theta T^*(\bu_\alpha^*,\bmu).
    \end{align}
    Then a key result in Lemma~\ref{lem:ratio} is established to transfer $\mathbb{E}_\mu[\tau_\delta]$ to $\mathbb{E}[f(\boldsymbol{c}, \boldsymbol{\mu};\tau_\delta)]$, which reveals the deep relationship between the cumulative cost and the sample complexity.

    \begin{lemma}\label{lem:ratio}
        Using the CAET algorithm with the optimal sampling proportion set as $\bu_\alpha^*=\bu_\alpha(\bc,\bmu)$, when $\delta\rightarrow 0$, the complexity $T^*(\boldsymbol{u}_\alpha^*,\boldsymbol{\mu})$ in Equation~\eqref{formula:upperbound} satisfies
        \begin{align*}
            \lim\limits_{\delta\rightarrow 0}T^*(\boldsymbol{u}_\alpha^*,\boldsymbol{\mu})^{-1}\dfrac{\mathbb{E}_\mu[\tau_\delta]}{\mathbb{E}[f(\boldsymbol{c}, \boldsymbol{\mu};\tau_\delta)]}=T^*(\boldsymbol{c},\boldsymbol{\mu})^{-1}.
        \end{align*}
    \end{lemma}
    Thus, it can be observed that for Bernoulli bandit $\boldsymbol{\mu}$, the choice $\beta(t,\delta)=\log(2tK(K-1)/\delta)$ in the stopping rule is $\delta$-PAC. Also, with the sampling rule given above, the stopping time $\tau_\delta$ is almost surely finite. When $\delta$ is small enough, its expectation approaches $T^*(\boldsymbol{\mu})\log(1/\delta)$. Thus, the optimal sample complexity is established in Theorem~\ref{the:lower-bound}. 
\end{Proofsketch}
More generally, for the exponential family bandits, combining Proposition~\ref{pro:Dev-Thre} and Theorem~\ref{thm:upperbound},
it can be obtained that for every $\theta >1$, there exists an exploration rate such that CAET is $\delta$-PAC and satisfies
\begin{align*}
    \limsup\limits_{\delta \rightarrow 0}\dfrac{\mathbb{E}[f(\boldsymbol{c}, \boldsymbol{\mu};\tau_\delta)]}{\log(1/\delta)}\leq \theta T^*(\boldsymbol{c},\boldsymbol{\mu}).
\end{align*}
Finally, we note that the $\delta$-dependency of the choice of $\alpha$ is related to the convergence rate of sample complexity and the order of stopping time. Especially, a faster convergence rate of $\alpha$ to $1$ improves the convergence rate of the algorithm, but at the expense of increasing the stopping time. The proposed CAET algorithm takes $\alpha=1-\log^{-r}(1/\delta)$ with $0<r<1/2$, achieving stopping time of order $O(\log^{1+r}(1/\delta))$.

\section{DISCUSSION ON SPECIAL CASES}\label{sec:regretmin}
The upper bound analyses in Theorem~\ref{thm:upperbound} are for the general pairwise pure exploration problem, which can be further refined for various specifically targeted tasks, e.g., best-arm identification and ranking identification. Details on these two specifications are articulated in Appendix~\ref{sec:ranking}. In this section, we provide additional results on extending the study to the problem of \emph{regret minimization}.

We define the optimal action $a^*$ as the arm with the highest expected reward, i.e., $a^*:=\argmax_{a\in\calA}\mu_a$. Also, the sub-optimal gap between arm $a$ and optimal arm $a^*$ is denoted as $\Delta_a:=\mu_{a^*}-\mu_a$.

To realize the regret minimization setting, we can track the sub-optimal gap through the difference of respective sample means in each round. In round $t$, we take $
    \boldsymbol{\nu}(t)=(0,\nu_{\Delta_2}(t),\nu_{\Delta_3}(t),\ldots ,\nu_{\Delta_K}(t))$,
where $\nu_{\Delta_i}(t)=\hat{\mu}_1(t)-\hat{\mu}_i(t)$ is the estimation of the sub-optimal gap. This scheme of observing the cost vector slightly differs from the setting described in Section~\ref{sec:problem} where the costs are always sampled from a certain distribution. However, as shown in the proofs, the proposed CAET algorithm is still functional in this scenario.

Furthermore, the discussions of BAI in Appendix~\ref{sec:ranking} demonstrate the commitment time of CAET to the optimal arm is of order $O(\log(1/\delta)^{1+r})$. Then, with $\delta = 1/T$ and an explore-then-commit (ETC) scheme (i.e., keep pulling the identified arm after commitment), CAET achieves
\begin{align}\label{for:reg1}
    R_{\bmu}(T)&\leq \theta T^*(\boldsymbol{\Delta},\bmu)\log(T),\\
    T^*(\boldsymbol{\Delta},\bmu)&=\sum\limits_{\Delta_a>0}\frac{\Delta_a}{\KL(\mu_a,\mu_1)},
\end{align}
which is further elaborated in Appendix~\ref{appendix:proveequ}.

Similar ideas of leveraging the ETC scheme to perform regret minimization using BAI designs have been recently studied in \cite{zhang2023fast}. For Bernoulli bandits, both CAET and \cite{zhang2023fast} achieve asymptotically optimal regret. For general exponential bandits, CAET is capable of achieving an asymptotically near-optimal performance (with $\theta>1$) without a pre-determined stopping time, which is required in \cite{zhang2023fast}. Note that although the commitment time $O(\log(1/\delta)^{1+r})$ of CAET is slightly larger than the pre-determined stopping time order $O(\log(1/\delta))$ in \cite{zhang2023fast}, we can take $r$ close enough to zero to approach their result.

\section{NUMERICAL EXPERIMENTS}
\label{sec:exp}
In this section, numerical experiment results are reported to evaluate the effectiveness and efficiency of the proposed CAET algorithm. The task of ranking identification is considered in the experiments with a focus on three-arm bandits, while the cost vector is set as the sub-optimal gap vector to also achieve the goal of regret minimization. A more explicit theoretical result for this particular task is provided in Proposition~\ref{pro:three-arm} in Appendix~\ref{sec:ranking}. Fig.~\ref{fig:group1} reports the cumulative regrets obtained by CAET over two Bernoulli bandit instances, $\bmu_1=(3, 4, 2), r=0.4, \theta=1.2$ and $\bmu_2=(1.4, 0.8, 0.3), r=0.4, \theta=1.2$ with truncation function $D_{\delta}(\cdot)$ taking $0.1\log^{-0.1}(1/\delta)$ as the threshold. The established theoretical lower and upper bounds are also plotted for comparison purposes. It can be observed that when $\delta \to 0$, the performance of CAET is indeed limited by the upper bound and also capable of approaching the lower bound, which corroborates the theoretically asymptotic optimality of CAET.

\begin{figure}[tbh]
    \setlength{\abovecaptionskip}{5pt}
    \centering
    \subfigure{
    \includegraphics[width=0.45 \textwidth]{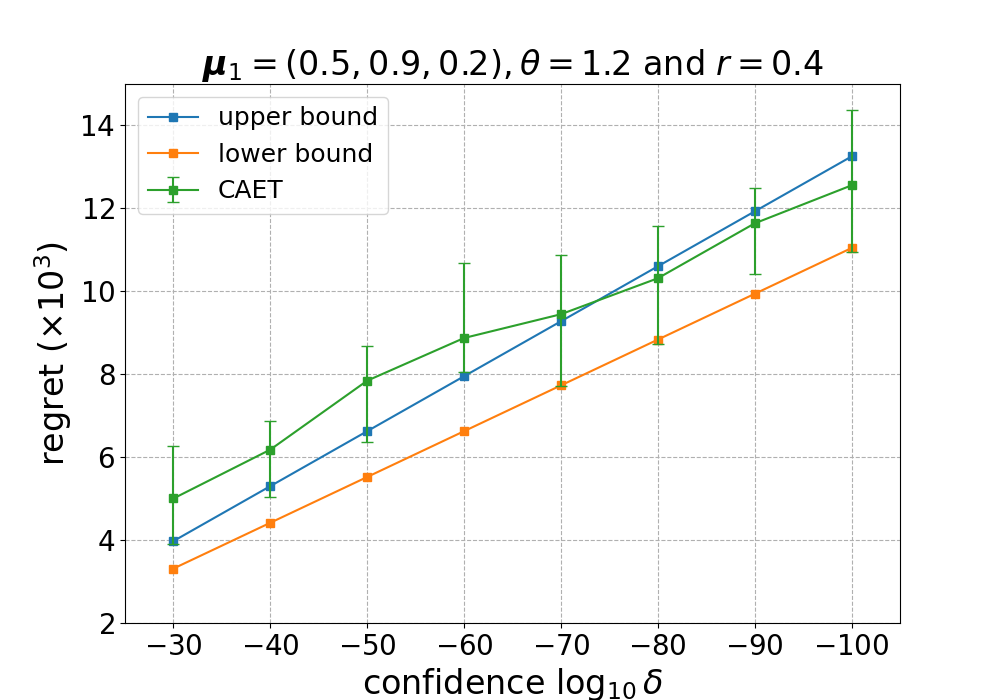}
    }
    \hspace{2mm}
    \subfigure{
    \includegraphics[width=0.45 \textwidth]{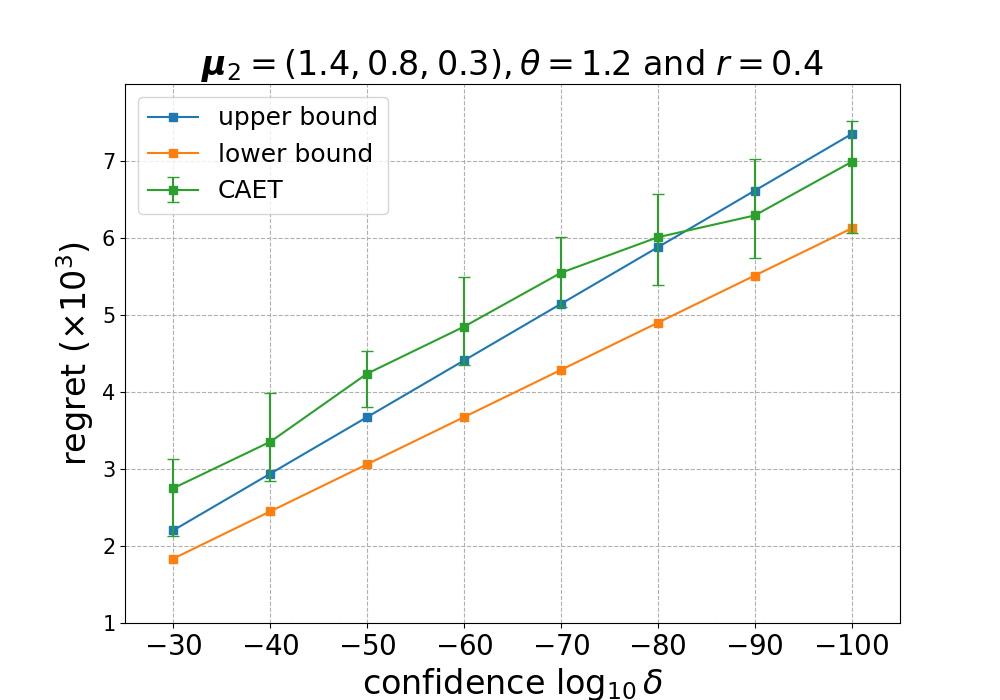}
    }
    \caption{Experimental results}
    \label{fig:group1}
    \vspace{-0.1in}
\end{figure}

\section{CONCLUSIONS}
\label{sec:conc}
This work introduced a general framework to study pure exploration problems in MAB. It focuses on identifying the pairwise relationships between arm pairs and can flexibly incorporate arm-specific costs to broaden the applicability. A performance lower bound was first established, which characterizes the fundamental limits of learning in the general framework. The novel CAET algorithm was then proposed, incorporating carefully crafted designs to handle arms associated with zero cost. Theoretical analyses demonstrated that CAET is capable of approaching the lower bound asymptotically, highlighting its optimality. An extension to regret minimization via the explore-then-commit scheme was further developed, which provably achieves the optimal performance asymptotically. Experimental results corroborated the effectiveness and efficiency of CAET.

To enhance the applicability of the CAET algorithm, one important future direction is to make it computationally more efficient due to the fact that a complex optimization problem needs to be solved in each round. One feasible solution is to consider a batched version of CAET, whose design and analysis are left for further investigation.

\section*{Acknowledgement}
The work of D. Wu, C. Shi and C. Shen was partially supported by the U.S. National Science Foundation (NSF) under awards ECCS-2033671, ECCS-2143559, ECCS-2029978, and CNS-2002902. The work of R. Zhou was supported in part by NSF grants 2139304,  2146838 and the Army Research Laboratory grant under Cooperative Agreement W911NF-17-2-0196.

\bibliography{ref}
\bibliographystyle{apalike}



\newpage

\onecolumn
 \thispagestyle{empty}

                \hsize\textwidth

    \linewidth\hsize \toptitlebar {\centering

    {\Large\bfseries Supplementary Material: \\Cost-Aware Optimal Pairwise Pure Exploration \par}}

    \bottomtitlebar

    \vspace{0.2in}
\appendix

\section{NOTATION}\label{app:notation}
To improve the readability, a notation table is provided. 
\begin{table}[h]
\centering
\begin{tabular}{|c|c|}
\hline
  $\calA$& The arm set\\
  $K$ & The number of arms \\
  $\boldsymbol{\mu}$& The mean vector of the arm set\\
  $\boldsymbol{c}$ & The cost vector\\
  $S_K$& The permutation group of $\calA$\\
  $\calG$ & The set of partition sets of the exploration task \\
  $\varphi$ & The answer function of instances\\
  $\sigma^{\boldsymbol{\mu}}$ & The index of descending order of $\bmu$ (see section~\ref{sec:definition})\\
  $\mathcal{B}_{ij}$ & The instances that the reward of arm-$j$ greater than arm-$i$ \\
  $\mathcal{I}_m$ & The indices set of some $(i,j)$ satisfying $\calG_m = \cap_{(i, j) \in \mathcal{I}_m} \mathcal{B}_{ij}$ \\ 
  $\delta$ & The confidence \\
  $N(\boldsymbol{c})$& The set of arms associated with zero cost\\
  $P(\boldsymbol{c})$& The set of arms associated with non-zero cost\\
  $\supp(\calI)$ & The support set of $\calI$ (see section~\ref{sec:notation})\\
  $N_{\calG}(\bc,\bmu)$ & The set of zero cost arm that in the support (see section~\ref{sec:notation})\\
  $P_{\calG}(\bc,\bmu)$ & The set of non-zero cost arm that in the support (see section~\ref{sec:notation})\\
  $\Alt(\boldsymbol{c},\boldsymbol{\mu})$ &  The alternative set of $\bc,\bmu$ (see formula (\ref{for:altalt}))\\
  $\Delta_K$ & The $K-1$ dimension probability simplex\\
  $\Sigma_{P_{\calG}(\bc,\bmu)}$& A subset of $\Delta_K$ (see formula (\ref{for:distributiondis}))\\
  $\boldsymbol{\omega}^*(\boldsymbol{c},\boldsymbol{\mu})$ & The optimal solution to the optimization problem\\
  $\boldsymbol{u}^*(\boldsymbol{c},\boldsymbol{\mu})$ & The optimal pulling ratio of the positive cost arms\\
  $\boldsymbol{u}_\alpha (\bc_\delta(t),\hat{\bmu}(t))$ & The optimal pulling ratio of all arms\\
  $\alpha$ & The pulling ratio of zero cost arms and non-zero cost arms\\
  $r$ & The parameter in $\alpha$\\
  $r'$ & The parameter in the threshold of truncation function $D_{\delta}$ \\
  $\gamma_0$ & The parameter in the threshold of truncation function $D_{\delta}$\\
    \hline
\end{tabular}
\end{table}

\section{DISCUSSIONS}
\subsection{Broad Impact}\label{subapp:impact}
This work introduces a general framework to study the pure-exploration problems in MAB, which broadens the scope of previous studies. With pure exploration being one of the core focuses in MAB and the theoretical nature of this work, we do not foresee major negative social impacts.

\subsection{Limitations and Future Works}\label{subapp:limitation}
This work introduces a general framework to study the pure exploration problem, with a focus on the pure pairwise exploration tasks. It would be an interesting direction to further investigate the framework to its full potential beyond the pairwise requirement, leveraging solely the given partitions. Also, this work mainly considers the fixed-confidence setting, while the dual fixed-budget setting \citep{audibert2010best,carpentier2016tight,komiyama2022minimax,wang2024best} is also an important topic worth future studies. Moreover, this work focuses on the basic tabular scenario without assuming any relationships between arms, while the established framework and the obtained insights may also inspire additional studies in other scenarios, e.g., linear bandits \citep{soare2014best,jedra2020optimal}.

\section{LOWER BOUND}
\subsection{Proof of Theorem~\ref{the:lower-bound} of the Main Paper}\label{app:lowerbound}
\begin{Proof}
Let $\varphi(\bmu)=m$ and exploration task be $\calG$. 
First, we recall the following lemma from \citet{kaufmann2016complexity}.
\begin{lemma}[Lemma 1 in \citet{kaufmann2016complexity}]\label{lem:change-dis}
    Let $v$ and $v'$ be two bandit models with $K$ arms such that for all $a\in [K]$, the distributions $v_a$ and $v_a'$ are mutually absolutely continuous. For any almost-surely finite stopping time $\sigma$ with respect to $(\calF_t)_t$, it holds that
    \begin{align*}
        \sum\limits_{a\in [K]}\mathbb{E}_v[N_a(\sigma)]\KL(v_a,v'_a)\geq \sup\limits_{\mathcal{E}\in \calF_\sigma} \hat{d}(\Pb_v(\mathcal{E}), \Pb_{v'}(\mathcal{E}))
    \end{align*}
    where $\hat{d}(x,y):=x \log(x/y)+(1-x) \log ((1-x)/(1-y))$ is the binary relative entropy, with the convention that $\hat{d}(0,0)=\hat{d}(1,1)=0$.
\end{lemma}
Recall the alternative set is:
\begin{align*}
    \Alt(\boldsymbol{c},\boldsymbol{\mu})=\{\boldsymbol{\lambda} \in S: \varphi(\boldsymbol{\lambda})\neq \varphi(\bmu), \lambda_i=\mu_i, \forall i \notin P_\calG(\boldsymbol{c},\bmu)\}. 
\end{align*}

To prove Theorem~\ref{the:lower-bound}, we can take $\calE=\{\varphi(\boldsymbol{\mu})=m\}$ in \Cref{lem:change-dis}, which leads to 
\begin{align*}
    \forall \boldsymbol{\lambda}\in \Alt(\boldsymbol{c},\boldsymbol{\mu}),\quad \sum\limits_{a\in \calA}d(\mu_a,\lambda_a)\Eb[N_a(\tau_\delta)] = \sum\limits_{a\in P_\calG(\boldsymbol{c},\bmu)}d(\mu_a,\lambda_a)\Eb[N_a(\tau_\delta)]\geq \kl(\delta,1-\delta),
\end{align*}
where the first equation is from the fact that $\mu_a=\lambda_a$ for $a\notin P_\calG(\boldsymbol{c},\bmu)$ according to the definition of $\Alt(\boldsymbol{c},\boldsymbol{\mu})$.

Thus, it can be obtained that
\begin{align*}
    \kl(\delta,1-\delta)&\leq \inf\limits_{\boldsymbol{\lambda}\in \Alt(\boldsymbol{c},\boldsymbol{\mu})}\mathbb{E}^{\calG}[f(\boldsymbol{c}, \boldsymbol{\mu};\tau_\delta)]\left(\sum\limits_{a\in P_\calG(\boldsymbol{c},\bmu)}\frac{c_a\mathbb{E}[N_a]}{\mathbb{E}^\calG[f(\boldsymbol{c}, \boldsymbol{\mu};\tau_\delta)]}\frac{d(\mu_a,\lambda_a)}{c_a}\right)\\
    &\leq \mathbb{E}^\calG[f(\boldsymbol{c}, \boldsymbol{\mu};\tau_\delta)]\sup\limits_{\omega \in \Sigma_{P_\calG(\boldsymbol{c},\bmu)}}\inf\limits_{\boldsymbol{\lambda}\in \Alt(\boldsymbol{c},\boldsymbol{\mu})}\left(\sum\limits_{a \in P_\calG(\boldsymbol{c},\bmu)} \omega_a \frac{d(\mu_a,\lambda_a)}{c_a}\right)
\end{align*}
where the second inequality leverages the definition that $\mathbb{E}^\calG[f(\boldsymbol{c}, \boldsymbol{\mu};\tau_\delta)]:=\sum_{a\in P_\calG(\boldsymbol{c},\bmu)}c_a\mathbb{E}[N_a(\tau_\delta)]$. We have
\begin{align*}
    \mathbb{E}^\calG[f(\boldsymbol{c}, \boldsymbol{\mu};\tau_\delta)]\geq T^*(\bc,\bmu)\kl(\delta,1-\delta),
\end{align*}
and $$\mathbb{E}[f(\boldsymbol{c}, \boldsymbol{\mu};\tau_\delta)]\geq \mathbb{E}^\calG[f(\boldsymbol{c}, \boldsymbol{\mu};\tau_\delta)].$$ Thus,
\begin{align*}
    \mathbb{E}[f(\boldsymbol{c}, \boldsymbol{\mu};\tau_\delta)]\geq T^*(\bc,\bmu)\kl(\delta,1-\delta),
\end{align*}
which concludes the proof.
\end{Proof}

\subsection{Characteristic Time and Optimal Proportions}\label{sec:character}
In the following, additional properties of the lower bound are further provided. In particular, for every $\alpha\in [0,1]$, we denote
\begin{align*}
    I_\alpha(\mu_1,\mu_2):=\alpha d(\mu_1,\alpha \mu_1+(1-\alpha)\mu_2)+(1-\alpha)d(\mu_2,\alpha \mu_1+(1-\alpha)\mu_2).
\end{align*}
Then, the following lemma can be established, which characterizes a more explicit form of the infimum and is helpful to compute $\boldsymbol{\omega}^*$ and $T^*$.
\begin{lemma}\label{lem:optgernal}
    For every $\omega \in \Sigma_{P_{\calG}(\bc,\bmu)}$, $\varphi(\bmu)=m$ and $\calG_m=\cap_{(i,j)\in\calI}\calB_{ij}$, it holds that
    \begin{align*}
        \inf_{\boldsymbol{\lambda}\in \Alt(\boldsymbol{c},\boldsymbol{\mu})}\left(\sum\limits_{a \in P_{\calG}(\bc,\bmu)} \omega_a \frac{d(\mu_a,\lambda_a)}{c_a}\right)&=\min\left[\min_{(a,b)\in \calI_m\atop a,b\notin N(\boldsymbol{c})}\left(\frac{\omega_b}{c_b}+\frac{\omega_a}{c_a}\right)I_{\frac{\omega_b/c_b}{\omega_b/c_b+\omega_a/c_a}}(\mu_b,\mu_a),\right.\\
        &\left.\min_{(a,b)\in \calI_m\atop a\notin N(\boldsymbol{c}), b\in N(\boldsymbol{c})} \left(\frac{\omega_a}{c_a} d(\mu_a,\mu_b)\right), \min\limits_{(a,b)\in \calI_m\atop b\notin N(\boldsymbol{c}),a\in N(\boldsymbol{c})}\left(\frac{\omega_b}{c_b} d(\mu_b,\mu_a)\right) \right].
    \end{align*}
\end{lemma}
\begin{Proof}
    Without loss of generality, we assume $\supp(\calI_m)=\calA$, and then $P_{\calG}(\bc,\bmu)=P(\bc)$ and $N_{\calG}(\bc,\bmu)=N(\bc)$.
    For $a>b,a\notin N(\bc),b\in N(\bc)$, we can first establish that
    \begin{align*}
        \inf\limits_{\boldsymbol{\lambda}\in S :\lambda_b=\mu_b\atop (\lambda_a-\lambda_b)(\mu_a-\mu_b)<0}\ (\sum\limits_{a\in P(\boldsymbol{c})} \frac{\omega_a}{c_a} d(\mu_a,\lambda_a))=\inf\limits_{\boldsymbol{\lambda}\in S :\lambda_b=\mu_b\atop (\lambda_a-\lambda_b)(\mu_a-\mu_b)\leq 0} \frac{\omega_a}{c_a} d(\mu_a,\lambda_a)=\frac{\omega_a}{c_a}d(\mu_a,\mu_b).
    \end{align*}
    Similarly, for $a>b,a\in N(\bc),b\notin N(\bc)$, we have
    \begin{align*}
        \inf\limits_{\boldsymbol{\lambda}\in S : \lambda_a=\mu_a\atop (\lambda_a-\lambda_b)(\mu_a-\mu_b)<0}\ (\sum\limits_{a\in P(\boldsymbol{c})} \frac{\omega_a}{c_a} d(\mu_a,\lambda_a))=\inf\limits_{\boldsymbol{\lambda}\in S :\lambda_a=\mu_a\atop (\lambda_a-\lambda_b)(\mu_a-\mu_b)\leq 0} \frac{\omega_b}{c_b} d(\mu_b,\lambda_b)=\frac{\omega_b}{c_b}d(\mu_b,\mu_a).
    \end{align*}
    Moreover, for $a>b,a\notin N(\bc),b\notin N(\bc)$, it holds that
    \begin{align*}
        \inf\limits_{\boldsymbol{\lambda}\in S:\atop (\lambda_a-\lambda_b)(\mu_a-\mu_b)<0}\ (\sum\limits_{a\in P(\boldsymbol{c})} \frac{\omega_a}{c_a} d(\mu_a,\lambda_a))=\inf\limits_{\boldsymbol{\lambda}\in S: \atop(\lambda_a-\lambda_b)(\mu_a-\mu_b)\leq 0} (\frac{\omega_b}{c_b} d(\mu_b,\lambda_b)+\frac{\omega_a}{c_a} d(\mu_a,\lambda_a)).
    \end{align*}
    Assuming $\mu_a<\mu_b$ without loss of generality, minimizing
    \begin{align*}
        f(\lambda_b,\lambda_a)=\frac{\omega_b}{c_b}d(\mu_b,\lambda_b)+\frac{\omega_a}{c_a}d(\mu_a,\lambda_a),
    \end{align*}
    is a convex optimization problem under the constraint $\lambda_a\geq \lambda_b$, which can be solved in closed form. In fact, the minimum is 
    \begin{align*}
        \lambda_b=\lambda_a=\frac{\omega_b/c_b}{\omega_b/c_b+\omega_a/c_a}\mu_b+\frac{\omega_a/c_a}{\omega_b/c_b+\omega_a/c_a}\mu_a
    \end{align*}
    and the optimal function value can be written as $(\frac{\omega_b}{c_b}+\frac{\omega_a}{c_a})I_{\frac{\omega_b/c_b}{\omega_b/c_b+\omega_a/c_a}}(\mu_b,\mu_a)$.    
    
    Combining the following Lemma~\ref{lem:decomp}, we have
    \begin{align*}
        & T^*(\boldsymbol{c},\boldsymbol{\mu})^{-1}=\sup\limits_{\omega \in \Sigma_{P(\boldsymbol{c})}}\min\limits_{(a,b)\in\calI_m}\inf\limits_{\blambda\in S:\lambda_i=\mu_i,\forall i\in N(\bc)\atop(\lambda_a-\lambda_b)(\mu_a-\mu_b)<0}(\sum\limits_{a \in P(\boldsymbol{c})} \omega_a \frac{d(\mu_a,\lambda_a)}{c_a})\\
        &=\sup\limits_{\omega \in \Sigma_{P(\boldsymbol{c})}}\min\bigg[\min\limits_{(a,b)\in \calI_m\atop a,b\notin N(\boldsymbol{c})}(\frac{\omega_b}{c_b}+\frac{\omega_a}{c_a})I_{\frac{\omega_b/c_b}{\omega_b/c_b+\omega_a/c_a}}(\mu_b,\mu_a),
        \min\limits_{(a,b)\in \calI_m\atop a\notin N(\boldsymbol{c}),b\in N(\boldsymbol{c})}(\frac{\omega_a}{c_a} d(\mu_a,\mu_b)), 
        \min\limits_{(a,b)\in \calI_m\atop b\notin N(\boldsymbol{c}),a\in N(\boldsymbol{c})}\ (\frac{\omega_b}{c_b} d(\mu_b,\mu_a))\bigg],
    \end{align*}
    which concludes the proof.
\end{Proof}


\begin{lemma}\label{lem:decomp}
    Let $\mu\in S$ such that $\bmu=(\mu_1,\ldots,\mu_K)$ and $\varphi(\bmu)=m$. We have
    \begin{align*}
        \Alt(\bc,\bmu)=\bigcup_{(a,b)\in \calI_m} \{\blambda\in S:(\lambda_a-\lambda_b)(\mu_a-\mu_b)<0,\ \lambda_i=\mu_i,\forall i\notin P_\calG(\bc,\bmu)\}.
    \end{align*}
\end{lemma}
\begin{Proof}
    Let $S_K$ be the permutation group of order $K$. Recall that
    \begin{align*}
        \calI_m\subseteq\calH=\{(a,b):1\leq b\leq a\leq K\}.
    \end{align*}
    Assume that $\varphi(\bmu)=m$ and $\calG_m=\cap_{(i,j)\in\calI_m}\calB_{ij}$. Consider set 
    \begin{align*}
        S(\bmu)=\bigcup\limits_{(a,b)\in\calI_m}\calB_{ba}.
    \end{align*}
    Assume $\Alt(\boldsymbol{\mu})=\{\boldsymbol{\lambda}\in S:\varphi(\boldsymbol{\lambda})\neq\varphi(\boldsymbol{\mu})\}$.
    If $\blambda\in \Alt(\bmu)$, there must exist $(i,j)\in \calI$, such that 
    $(\sigma^{\blambda}_i-\sigma^{\blambda}_j)(\sigma^{\bmu}_i-\sigma^{\bmu}_j)<0$.
    Thus
    \begin{align*}
        \Alt(\bmu)\subset S(\bmu).
    \end{align*}
    Also, it is obvious that
    \begin{align*}
        \Alt(\bmu)\supset S(\bmu).
    \end{align*}
    We thus have 
    \begin{align*}
        \Alt(\bmu)=S(\bmu)=\bigcup\limits_{(a,b)\in\calI_m}\calB_{ba}=\bigcup_{(a,b)\in \calI_m} \{\blambda\in S:(\lambda_a-\lambda_b)(\mu_a-\mu_b)<0\}.
    \end{align*}
    After adding cost vector $\bc$ and $\supp(\calI(\bmu))$, we can get
    \begin{align*}
        \Alt(\bc,\bmu)=\bigcup_{(a,b)\in \calI_m} \{\blambda\in S:(\lambda_a-\lambda_b)(\mu_a-\mu_b)<0,\ \lambda_i=\mu_i,\forall i\notin P_\calG(\bc,\bmu)\},
    \end{align*}
    which concludes the proof.
\end{Proof}

\section{TRACKING RESULTS}\label{appendix:track}
Here, we discuss the properties of our sampling and algorithm. Recall we track the proportion of 
\begin{align*}
    \boldsymbol{u}_\alpha (\bc_\delta(t),\hat{\bmu}(t))= \underbrace{\lv_\alpha(\bc_\delta(t),\hat{\bmu}(t))}_{\text{arms in $N_{\calG}(\bc_\delta(t),\hat{\bmu}(t))$}}+\underbrace{(1-\alpha)\boldsymbol{u}^*(\bc_\delta(t),\hat{\bmu}(t))}_{\text{arms in $P_\calG(\bc_\delta(t),\hat{\bmu}(t))$}},
\end{align*}
and when $N_{\calG}(\bc_\delta(t),\hat{\bmu}(t))=\emptyset$, our track proportion is directly set as $\boldsymbol{u}^*(\bc_\delta(t),\hat{\bmu}(t))$.

The proposed algorithm CAET uses a specific sample distribution $\bu_{\alpha}(\bc,\bmu)=(u_1,\ldots,u_K), \alpha=\alpha(\delta)$. The following properties (Lemma~\ref{lem:cumulative-time}, Proposition~\ref{pro:time-converge}) can be obtained with their proofs extended from Lemma 7 and Proposition 9 in \cite{garivier2016optimal}.  
\begin{lemma}\label{lem:cumulative-time}
    For all $t\geq 1$ and arm $a\in \calA$. For any fixed $\delta>0$ the tracking strategy with sample distribution $\bu_{\alpha}(\bc,\bmu)=(u_1,\ldots,u_K)$, ensures that $N_a(t)\geq \sqrt{t+K^2}-2K$ and 
    \begin{align*}
        \max\limits_{1\leq a \leq K}\left|N_a(t)-\sum\limits_{s=0}^{t-1}\bu_{a}(\bc_\delta(s),\hat{\bmu}(s))\right|\leq K(1+\sqrt{t}).
    \end{align*}
\end{lemma}
When $N_\calG(\bc,\bmu)= \emptyset$, our sampling strategy does not change with $\delta$, and each arm in $\supp(\calI(\bmu))$ has a positive cost. 
\begin{proposition}\label{pro:time-converge}
    When $N_\calG(\bc,\bmu)= \emptyset$, for all arm $a\in\supp(\calI(\bmu))$, the tracking strategy satisfies that
    \begin{align*}
        P\left(\lim\limits_{t\rightarrow \infty}\frac{N_a(t)}{t}=u_a\right)=1.
    \end{align*}
\end{proposition}

However, when $N_\calG(\bc,\bmu)\neq \emptyset$, recall that $\boldsymbol{u}_\alpha^*=(u_1,\ldots,u_K)=\lv_\alpha(\bc,\bmu)+\bu^*(\bc,\bmu)$ and the optimal proportion $u_a$ under $\delta$ is a function of $\delta$, $u_a=u_a(\delta)$, which is no longer a constant and thus means that Proposition~\ref{pro:time-converge} is not sufficient. We also denote $\bu^*(\bc,\bmu)=(u^*_1,\ldots,u^*_K)$. 

We need a proposition stronger than Proposition~\ref{pro:time-converge}. Before proving Proposition~\ref{pro:deltaconverge}, we introduce some lemma first.
\begin{lemma}\label{lem:event}
    Let $\boldsymbol{\nu}=(\nu_1,\ldots,\nu_K)$ be a subgaussian random vector with $\Eb(\nu_i)=\mu_i$. $X_{a,t},\ t=1,2,\ldots$, be a series i.i.d. sample from distribution $\nu_a$. Define event 
    \begin{align*}
        \mathcal{E}_M : = \left\{ \max_{a}\max \left( X_{a,1}, \frac{1}{2} (X_{a,1} + X_{a,2}), \ldots \right) \leq  M \right\}
    \end{align*}
    for sufficiently large $M$. We have $\mathbb{P}( \mathcal{E}_M^c ) \rightarrow 0$, as $M \rightarrow \infty$. 
\end{lemma}
\begin{Proof}
    Denote $\bar{\mu}_a(t)=\frac{1}{t}(X_{a,1}+\cdots+X_{a,t})$ and assume $\nu_i$ are $\sigma_i$-subgaussian random variable. Let 
    \begin{align*}
        \mu_{max}=\max \{\mu_1,\ldots,\mu_K\}\quad\textit{and}\quad \sigma_{max}=\max\{\sigma_1,\ldots,\sigma_K\}.
    \end{align*}
    By the operation of subgaussian random variables, we have $\bar{\mu}_a(t)$ is $\sigma_i/\sqrt{t}$-subgaussian which means, for $M>\mu_{max}$, we have
    \begin{align*}
        \Pb(\bar{\mu}_a(t)> M)\leq\exp{\left(-\frac{t(M-\mu_a)^2}{2\sigma_a^2}\right)}\leq\exp{\left(-\frac{t(M-\mu_{max})^2}{2\sigma_{max}^2}\right)}
    \end{align*}
    Thus, we have an union bound for $\Pb(\calE_M)$: 
    \begin{align*}
        \Pb(\calE_M)\leq \sum_{a} \sum_{t\geq 1} \mathbb{P}( \bar{\mu}_a(t) > M) \leq K \sum_{t \geq 1} \exp{\left(-\frac{t(M-\mu_{max})^2}{2\sigma_{max}^2}\right)} \leq K\frac{\exp{\left(-\frac{(M-\mu_{max})^2}{2\sigma_{max}^2}\right)}}{1-\exp{\left(-\frac{(M-\mu_{max})^2}{2\sigma_{max}^2}\right)}}
    \end{align*}
    and 
    \begin{align*}
        \lim\limits_{M\rightarrow\infty}\exp{\left(-\frac{(M-\mu_{max})^2}{2\sigma_{max}^2}\right)}=0
    \end{align*}
    leads to our result: $\Pb(\calE^c_M)\rightarrow 0$ when $M\rightarrow\infty$.    
\end{Proof}

\begin{lemma}[Continuity]\label{lem:continuity}
    For $\bmu$ belongs to exactly one $\calG_m\in\calG$ and cost vector $\bc$, there exists a constant $\delta_0$ that when $\delta<\delta_0$, all of $\boldsymbol{\omega}^*(\bc,\bmu)$, $\bu^*(\bc,\bmu)$ and $\bu_\alpha(\bc,\bmu)$ are continuous at $(\bc,\bmu)$.
\end{lemma}
\begin{Proof}
    Recall $D_\delta([a_1,\ldots,a_K])=[b_1,\ldots,b_K]$ with $b_i=a_i$ if $a_i>\gamma_0\log^{-r'}(1/\delta)$ and $b_i=0$ if $a_i\leq\gamma_0\log^{-r'}(1/\delta)$. We take $\delta_0$ such that $\gamma_0\log^{-r'}(1/\delta)<\min_{i:c_i>0}\{c_i\}$. When $\delta\leq\delta_0$, we have $D_\delta(\bc)=\bc$.
    
    Since $\bu^*(\bc,\bmu)=G_{\bc}(\boldsymbol{\omega}^*(\bc,\bmu))$ and $\bu_\alpha(\bc,\bmu)=\lv(\bc,\bmu)+\boldsymbol{\omega}^*(\bc,\bmu)$, we only need to prove the continuity of $\lv(\bc,\bmu),\boldsymbol{\omega}^*(\bc,\bmu)$. 
    Recall that 
    \begin{align}\label{equ:111}
        & \boldsymbol{\omega}^*(\boldsymbol{c},\boldsymbol{\mu}):=\argmax\limits_{\boldsymbol{\omega} \in \Sigma_{P_{\calG}(\bc,\bmu)}}\inf\limits_{\boldsymbol{\lambda}\in \Alt(\boldsymbol{c},\boldsymbol{\mu})}\left\{\sum\limits_{a \in P_{\calG}(\bc,\bmu)} \omega_a \frac{d(\mu_a,\lambda_a)}{c_a}\right\}\nonumber\\
        &=\argmax\limits_{\boldsymbol{\omega} \in \Sigma_{P_{\calG}(\bc,\bmu)}} \min\bigg[\min_{(a,b)\in \calI_m\atop a,b\notin N(\boldsymbol{c})}\left(\frac{\omega_b}{c_b}+\frac{\omega_a}{c_a}\right)I_{\frac{\omega_b/c_b}{\omega_b/c_b+\omega_a/c_a}}(\mu_b,\mu_a),\min_{(a,b)\in \calI_m\atop {a\notin N(\boldsymbol{c})\atop b\in N(\boldsymbol{c})}} \left(\frac{\omega_a}{c_a} d(\mu_a,\mu_b)\right), \min\limits_{(a,b)\in \calI_m\atop {b\notin N(\boldsymbol{c})\atop a\in N(\boldsymbol{c})}}\left(\frac{\omega_b}{c_b} d(\mu_b,\mu_a)\right) \bigg]
    \end{align}
    and 
    \begin{align*}
        P_\calG(\bc,\bmu)=P(\bc)\cap\supp(\calI(\bmu))=\left\{a:a\in\calA, c_a>0, a\in\supp(\calI(\bmu))\right\}.
    \end{align*}
    Taking $\varepsilon_0=\min\{\min_i\{c_i-\gamma_0\log^{-r'}(1/\delta)\},\gamma_0\log^{-r'}(1/\delta)\}$ (note that $\delta<\delta_0$), when $|\bc'-\bc|<\varepsilon_0$, we have $N(D_\delta(\bc'))=N(\bc)$ and $P(D_\delta(\bc'))=P(\bc)$.

    For $\calG$, because $\bmu$ belongs to exactly one partition $\calG_m$, Letting 
    \begin{align*}
        \varepsilon_1=\frac{1}{2}\min\limits_{1\leq i,j\leq K\atop \mu_i\neq\mu_j}|\mu_i-\mu_j|
    \end{align*}
    when $|\bmu'-\bmu|<\varepsilon_1$, from the definition of $\varepsilon_1$, we can have $\varphi(\bmu')=\varphi(\bmu)$.

    Form the above discussion, for all $(\bc',\bmu')$ satisfying $|\bmu'-\bmu|<\varepsilon_1$ and $|\bc'-\bc|<\varepsilon_0$, there is $P_\calG(\bc',\bmu')=P_\calG(\bc,\bmu)$ and we can discuss the continuity in this neighborhood of $(\bc,\bmu)$ named it as $\mathcal{N}$. In this neighborhood, $\calI(\bmu')\equiv\cal(\bmu)$ and $N(D_\delta(c'))\equiv N(\bc)$, which suggests that $\boldsymbol{\omega}^*(\bc,\bmu)$ is made of some continuous function and some math operation ($\min$, $\argmax$) by Equation~\eqref{equ:111}. 
    Thus, $\boldsymbol{\omega}^*(\bc,\bmu)$ is continuous at $(\bc,\bmu)$ when $\delta<\delta_0$. 
    For $\lv_\alpha(\bc,\bmu)$, since in neighborhood $\mathcal{N}$, $P_\calG(\bc',\bmu')=P_\calG(\bc,\bmu)$, so $\lv_\alpha(\bc,\bmu)=\lv_\alpha(\bc',\bmu')$ for $(\bc',\bmu')\in\mathcal{N}$. Thus, $\lv_\alpha(\bc,\bmu)$ is also continuous and we conclude the proof.    
\end{Proof}

\begin{definition}
    For bandits model $\bmu=(\mu_1,\ldots,\mu_K)$, we call a sampling rule $\mathcal{R}$ \textbf{stable} if there exist constants $0<C_1$ and $C_2<1$ that
    \begin{align*}
        \lim\limits_{t\rightarrow \infty}\Pb\left(C_1<\frac{N_a(t)}{t}<C_2\right)=1,\quad \forall a=1,\ldots,K.
    \end{align*}
\end{definition}
Recall the stopping rule of CAET algorithm
\begin{align}\label{for:stoppingrule2}
    \tau_\delta =\inf\{t\in N: \exists\calG_m, \forall (a,b)\in\calI_m, Z_{a,b}(t)>\beta(t,\delta)\},
\end{align}
and we now present some properties regarding $\tau_\delta$.
\begin{proposition}
    For bandits model $\bmu=(\mu_1,\ldots,\mu_K)$, any stable sampling rule will ensure an almost sure finite stopping time $\tau_\delta$ under the stopping rule given in \Cref{for:stoppingrule2} with  $\beta(t,\delta)=\log(Ct^\theta/\delta), \theta\in[1,e/2]$, i.e.,
    \begin{align*}
            \Pb(\tau_\delta<\infty)=1.
    \end{align*}
\end{proposition}
\begin{Proof}
    Since the exploration task is pairwise, and contains only finite pairwise tests, we only need to consider the two-arm situation and Proposition~\ref{pro:stable} is enough.
\end{Proof}
\begin{proposition}\label{pro:stable}
    For the two-arm bandit model $\bmu=(\mu_1,\mu_2)$, any stable sampling rule will ensure an almost sure finite stopping time $\tau_\delta$ under the stopping rule given in \Cref{for:stoppingrule2} with  $\beta(t,\delta)=\log(Ct^\theta/\delta), \theta\in[1,e/2]$, i.e.,
    \begin{align*}
            \Pb(\tau_\delta<\infty)=1.
    \end{align*}
\end{proposition}
\begin{Proof}
    Recall when $\hat\mu_1(t)>\hat\mu_2(t)$,
    \begin{align*}
        Z_{1,2}(t)=N_1(t)d(\hat{\mu}_1(t),\hat{\mu}_{1,2}(t))+N_2(t)d(\hat{\mu}_2(t),\hat{\mu}_{1,2}(t)).
    \end{align*}
    Let $\calE=\{\forall a\in\{1,2\}, \hat\mu_a(t)\mathop{\rightarrow}\limits_{t\rightarrow\infty}\mu_a\}$. From the assumption on the sampling strategy and the Law of Large Numbers, $\Pb(\calE)=1$. When $t$ goes to infinity, both $d(\hat{\mu}_1(t),\hat{\mu}_{1,2}(t))$ and $d(\hat{\mu}_2(t),\hat{\mu}_{1,2}(t))$ are lower bounded by some constant $K(\bmu)$ where
    \begin{align*}
        K(\bmu)=\min\big\{d(\mu_1,C_2\mu_1+(1-C_2)\mu_2),d(\mu_2,C_2\mu_2+(1-C_2)\mu_1)\big\}.
    \end{align*}
    Therefore, for all $\epsilon>0$ there exists $t_1$ such that for all $t>t_1$
    \begin{align*}
        &d(\hat{\mu}_1(t),\hat{\mu}_{1,2}(t))>\frac{1}{1+\epsilon}K(\bmu),\\
        &d(\hat{\mu}_2(t),\hat{\mu}_{1,2}(t))>\frac{1}{1+\epsilon}K(\bmu).
    \end{align*}
    Hence, for $t>t_1$,
    \begin{align*}
        Z_{1,2}(t)>2tC_1\frac{1}{1+\epsilon}K(\bmu).
    \end{align*}
    Consequently,
    \begin{align*}
        \tau_\delta=\inf\{t\in\Nb:Z_{1,2}(t)\geq \beta(t,\delta)\} \leq t_1\vee \inf\{t\in\Nb:2tC_1(1+\epsilon)^{-1}K(\bmu)\geq \log(Ct^\alpha/\delta)\}.
    \end{align*}
    Using Lemma~\ref{lem:twoconstant} below, it follows that on $\calE$, as $\alpha\in[1,e/2]$,
    \begin{align*}
        \tau_\delta\leq t_1\vee \alpha(1+\epsilon)C_1^{-1}K(\bmu)^{-1}\left[\log\left(\frac{Ce(1+\epsilon)^\alpha}{\delta(C_1K(\bmu))^\alpha}\right)+\log{\log\left(\frac{C(1+\epsilon)^\alpha}{\delta(C_1K(\bmu))^\alpha}\right)}\right].
    \end{align*}
    Thus $\tau_\delta$ is finite on $\calE$, which concludes the proof.
\end{Proof}
\begin{lemma}\label{lem:twoconstant}
    For every $\alpha\in[1,e/2]$, for any two constant $\kappa_1,\kappa_2>0$, there exists
    \begin{align*}
        x=\frac{\alpha}{\kappa_1}\left[\log\left(\frac{\kappa_2e}{\kappa_1^\alpha}\right)+\log{\log\left(\frac{\kappa_2}{\kappa_1^\alpha}\right)}\right]
    \end{align*}
    such that $\kappa_1x\geq \log(\kappa_2x^\alpha)$.
\end{lemma}
\begin{Proof}
    This lemma has been stated in \cite{garivier2016optimal} which can be checked directly. It can also be seen as a by-product of well-known bounds on the Lambert W function.
\end{Proof}

\begin{lemma}\label{lem:as-cumulativetime}
    Let $\theta\in[1,e/2]$ and assume $\supp(\calI(\bmu))=\calA$. For any exploration task $\calG$ in bandit model $\bmu$, denote $\{\calR_\delta\}$ as a series of stable sampling strategies. Each sampling rule $\calR_\delta$, with the stopping rule given in \Cref{for:stoppingrule2} under threshold $\beta(t,\delta)=\log(Ct^\theta/\delta)$, has a stopping time $\tau_\delta=\tau_\delta(\calR_\delta)$  and respective pulling time $N_a^\delta(\tau_\delta)$. Then $\{\tau_\delta\}_\delta$ is a sequence of random variables that almost surely satisfy
    \begin{align*}
        \Pb\left(\liminf\limits_{\delta\rightarrow 0}\frac{N_a^\delta(\tau_\delta)}{\log(1/\delta)}>K(\bmu)\right)=1,\qquad a\in\calA,
    \end{align*}
    where $K(\bmu)$ is a constant of $\bmu$.
\end{lemma}

\begin{Proof}
    Define event $\Omega_0=\{\bar{\mu}_a(t)\mathop{\rightarrow}\limits_{t\rightarrow\infty}\mu_a, \forall a\}\cap \{\bar{c}_a(t)\mathop{\rightarrow}\limits_{t\rightarrow\infty}c_a, \forall a\}$. From the Law of Large Numbers, $\Pb(\Omega_0)=1$. Considering the event $\calE_m\cap\Omega_0$, we will prove the convergence conditioned on this event and then let $M$ go to infinity to conclude the proof (we use the fact of Lemma~\ref{lem:event}: $\lim_{M\rightarrow\infty}\Pb(\calE_M)=1$).

    Recall the stopping rule is 
    \begin{align*}
    \tau_\delta &= \inf\{t\in N: \exists\calG_m, \forall (a,b)\in\calI_m, Z_{a,b}(t)>\beta(t,\delta)\},
\end{align*}
    and when $\hat\mu_a(t)>\hat\mu_b(t)$,
    \begin{align*}
        Z_{a,b}(t)=N_a(t)d(\hat{\mu}_a(t),\hat{\mu}_{a,b}(t))+N_b(t)d(\hat{\mu}_b(t),\hat{\mu}_{a,b}(t))
    \end{align*}
    where
    \begin{align*}
        \hat{\mu}_{a,b}(t)=\frac{N_a(t)}{N_a(t)+N_b(t)}\hat{\mu}_a(t)+\frac{N_b(t)}{N_a(t)+N_b(t)}\hat{\mu}_b(t).
    \end{align*}
    First, we prove $\Pb(\lim\limits_{\delta\rightarrow 0}\tau_\delta=\infty)=1$:
    \begin{align*}
        \log(C/\delta)<&\log(C{\tau_\delta}^\theta/\delta)\\
        <&N_a^\delta(\tau_\delta)d(\hat{\mu}_a(\tau_\delta),\hat{\mu}_{a,b}(\tau_\delta))+N^\delta_b(\tau_\delta)d(\hat{\mu}_b(\tau_\delta),\hat{\mu}_{a,b}(\tau_\delta))\\
        <&\tau_\delta[d(\hat{\mu}_a(\tau_\delta),\hat{\mu}_{a,b}(\tau_\delta))+d(\hat{\mu}_b(\tau_\delta),\hat{\mu}_{a,b}(\tau_\delta))].
    \end{align*}
    Since conditioned on event $\calE_M$, both $d(\hat{\mu}_a(\tau_\delta),\hat{\mu}_{a,b}(\tau_\delta))$ and $d(\hat{\mu}_b(\tau_\delta),\hat{\mu}_{a,b}(\tau_\delta))$ are bounded, we get
    \begin{align*}
        \Pb(\lim\limits_{\delta\rightarrow 0}\tau_\delta=\infty|\calE_M)=1
    \end{align*}
    and let $M$ approach to infinity, by Lemma~\ref{lem:event}, we can get the desire result:
    \begin{align*}
        \Pb(\lim\limits_{\delta\rightarrow 0}\tau_\delta=\infty)=1.
    \end{align*}
    
    For any $b\in\supp(\calI(\bmu))=\calA$, there exists $a$, such that $(a,b)\in\calI(\bmu)$ or $(b,a)\in\calI(\bmu)$. We assume $(a,b)\in\calI(\bmu)$. Then, $\tau_\delta$ satisfies
    \begin{align*}
        N_a^\delta(\tau_\delta)d(\hat{\mu}_a(\tau_\delta),\hat{\mu}_{a,b}(\tau_\delta))+N_b^\delta(\tau_\delta)d(\hat{\mu}_b(\tau_\delta),\hat{\mu}_{a,b}(\tau_\delta))>\log(C{\tau_\delta}^\theta/\delta).
    \end{align*}
    Assuming $N^\delta_a(\tau_\delta)>N^\delta_b(\tau_\delta)$, we first consider the following limits
    \begin{align*}
        \lim\limits_{t=x/y\rightarrow\infty}x\cdot d\left(\mu_a,\frac{x\mu_a+y\mu_b}{x+y}\right)\bigg/y=\lim\limits_{t\rightarrow\infty}\frac{ d\left(\mu_a,\frac{t\mu_a+\mu_b}{t+1}\right)}{1/t}.
    \end{align*}
    By the L'Hospital rule, we only need to calculate
    \begin{align*}
        &\lim\limits_{t\rightarrow\infty}\frac{\frac{d}{dt} d\left(\mu_a,\frac{t\mu_a+\mu_b}{t+1}\right)}{-1/t^2}=\lim\limits_{t\rightarrow\infty}-t^2\frac{d}{dt} d\left(\mu_a,\frac{t\mu_a+\mu_b}{t+1}\right)\\
        =&\lim\limits_{t\rightarrow\infty}-t^2\frac{(\mu_b-\mu_a)}{t+1}\frac{1}{b''(b^{-1}(y_0))}\frac{(\mu_a-\mu_b)}{(t+1)^2}\\
        =&\lim\limits_{t\rightarrow\infty}\frac{(\mu_a-\mu_b)^2}{b''(b^{-1}(y_0))}\frac{t^2}{(t+1)^3}\\
        =&0
    \end{align*}
    where $\frac{d}{dx}d(x,y)=(y-x)/b''(b^{-1}(y))$ and 
    \begin{align*}
        y_0=\frac{t\mu_a+\mu_b}{t+1}\in [\mu_a,\mu_b]
    \end{align*}
    is bounded.
    So, we have 
    \begin{align*}
        \lim\limits_{t=x/y\rightarrow\infty}x\cdot d\left(\mu_1,\frac{x\mu_a+y\mu_b}{x+y}\right)\bigg/y=0
    \end{align*}
    which shows there exists $N_0\in\Nb$ such that when $x/y>N_0$, we have
    \begin{align*}
        x\cdot d\left(\mu_a,\frac{x\mu_a+y\mu_b}{x+y}\right)<y.
    \end{align*}
    Now, we continue our proof. If $N_a^\delta(\tau_\delta)/N_b^\delta(\tau_\delta)>N_0$, we have  
    \begin{align*}
        N_a^\delta(\tau_\delta)d(\hat{\mu}_a(\tau_\delta),\hat{\mu}_{a,b}(\tau_\delta))<N_b^\delta(\tau_\delta).
    \end{align*}
    Thus,
    \begin{align*}
        \log(C/\delta)<&\log(C{\tau_\delta}^\theta/\delta)\\
        <&N_a^\delta(\tau_\delta)d(\hat{\mu}_a(\tau_\delta),\hat{\mu}_{a,b}(\tau_\delta))+N^\delta_b(\tau_\delta)d(\hat{\mu}_b(\tau_\delta),\hat{\mu}_{a,b}(\tau_\delta))\\
        <&N_b^\delta(\tau_\delta) + N_b^\delta(\tau_\delta)d(\hat{\mu}_b(\tau_\delta),\hat{\mu}_{a,b}(\tau_\delta))\\
        <&N^\delta_b(\tau_\delta)(1+d(\hat{\mu}_b(\tau_\delta),\hat{\mu}_{a,b}(\tau_\delta))).
    \end{align*}
    When $N_a^\delta(\tau_\delta)/N_b^\delta(\tau_\delta)<N_0$, we have 
    \begin{align*}
        \log(C/\delta)<&\log(C{\tau_\delta}^\theta/\delta)\\
        <&N_a^\delta(\tau_\delta)d(\hat{\mu}_a(\tau_\delta),\hat{\mu}_{a,b}(\tau_\delta))+N_b^\delta(\tau_\delta)d(\hat{\mu}_b(\tau_\delta),\hat{\mu}_{a,b}(\tau_\delta))\\
        <&N_b^\delta(\tau_\delta)\big[N_0d(\hat{\mu}_a(\tau_\delta),\hat{\mu}_{a,b}(\tau_\delta))+d(\hat{\mu}_b(\tau_\delta),\hat{\mu}_{a,b}(\tau_\delta))\big].
    \end{align*}
    Thus
    \begin{align}\label{equ:bound}
        \log(C/\delta)<N_b^\delta(\tau_\delta)\cdot\max\big\{&(1+d(\hat{\mu}_b(\tau_\delta),\hat{\mu}_{a,b}(\tau_\delta))),\nonumber \\
        &(N_0d(\hat{\mu}_a(\tau_\delta),\hat{\mu}_{a,b}(\tau_\delta))+d(\hat{\mu}_b(\tau_\delta),\hat{\mu}_{a,b}(\tau_\delta)))\big\}.
    \end{align}
    Since $\Pb(\lim\limits_{\delta\rightarrow 0}\tau_\delta=\infty)=1$ and Lemma~\ref{lem:cumulative-time} ensures that $N^\delta(\tau_\delta)\geq \sqrt{\tau_\delta+K^2}-2K$, we can have that, for all arm $a\in\calA$,
    \begin{align*}
        \Pb(\liminf\limits_{\delta\rightarrow 0}N_a^\delta(\tau_\delta)=\infty)=1,\quad a\in\calA
    \end{align*}
    Therefore, by the Law of Large Numbers, both $d(\hat{\mu}_a(\tau_\delta),\hat{\mu}_{a,b}(\tau_\delta))$ and $d(\hat{\mu}_b(\tau_\delta),\hat{\mu}_{a,b}(\tau_\delta))$ are bounded with probability $1$. Thus, by Equation~\eqref{equ:bound}, we have
    \begin{align*}
        \Pb\left(\liminf\limits_{\delta\rightarrow 0}\frac{N_b^\delta(\tau_\delta)}{\log(1/\delta)}>K(\bmu)\right)=1
    \end{align*}
    for some constant $K(\bmu)$. Repeating the same process for the other arms, we have 
    \begin{align*}
        \Pb\left(\liminf\limits_{\delta\rightarrow 0}\frac{N_a^\delta(\tau_\delta)}{\log(1/\delta)}>\hat{K}(\bmu)\right)=1,\quad a\in\calA, 
    \end{align*}
    which concludes the proof.    
\end{Proof}

Before proving the main proposition, we introduce a concentration lemma to handle the bias of the sample means during adaptive sampling and adaptive stopping.
\begin{lemma}
    Consider an adaptive sampling algorithm and stopping rule. For a fixed arm $k\in \calA$ with a finite $2p$-norm, where  $p>1$, and any random time $\tau$ such that $N_k(\tau)\geq 3$ almost surely, it holds that, for any $\varsigma>0$,
    \begin{align*}
        \Pb\left(\frac{N_k(\tau)}{\log{N_k(\tau)}}\left(\frac{\hat{\mu}_k(\tau)-\mu_k}{q_k}\right)^2\geq \varsigma \right)\leq \frac{C_p}{\varsigma^p}
    \end{align*}
    where $C_p$ is a constant depending only on $p$, and $q_k$ is a constant depending only on $\mu_k$.
\end{lemma}
The proof of this lemma has been given in Lemma 4.4 in \cite{shin2019bias}. For any $\delta>0$, our sampling rule $\calR_\delta$ satisfies $N_k(t)\geq \sqrt{t+K^2}-2K,\ \ \forall k\in\calA$. Thus, when $\tau=\tau'_\delta$ is large enough, we have 
\begin{align*}
    \frac{N_k(\tau_\delta')}{\log N_k(\tau_\delta')}> N_k(\tau_\delta')^{2/3}>\tau_\delta'^{1/4},
\end{align*}
and
\begin{align*}
    \Pb\left({\tau'_\delta}^{1/4}\left(\frac{\hat{\mu}_k(\tau'_\delta)-\mu_k}{q_k}\right)^2\geq \varsigma \right)\leq \frac{C_p}{\varsigma^p}.
\end{align*}
We take $\tau'_\delta=\log^{r_1}(1/\delta), r_1< 1$. When $\delta$ is small enough, we will have $\tau'_\delta\leq\tau_\delta$. 
Thus, we have
\begin{align}\label{equ:concen}
    \Pb\left(\hat{\mu}_k(\tau'_\delta)-\mu_k\geq q_k\varsigma^{1/2}(\tau'_\delta)^{-1/8} \right)\leq \frac{C_p}{\varsigma^p}.
\end{align}
The above discussion leads to the following lemma:
\begin{proposition}\label{pro:bias-stop}
    For the series of sampling rules $\{\calR_\delta \}_\delta$ in the CAET algorithm, we can precisely determine the arm with zero cost before a time on the order of $\tau_\delta'=\log^{r_1}(1/\delta)=o(\log(1/\delta))=o(\tau_\delta(1-\alpha))$, i.e.,
    \begin{align*}
        \Pb\left(\lim\limits_{\delta\rightarrow 0}N(\bc_\delta(\tau'_\delta))=N(\bc)\right)=1.
    \end{align*}
\end{proposition}
\begin{Proof}
    Recall $\bc_\delta(t)=D_\delta(\hat{\bc}(t))$ and $D_\delta(\cdot)$ is a truncation function with threshold $\gamma_0\log^{-r'}(1/\delta)$. When we take $\varsigma=\gamma_0^2\log^{r_1/4-2r'}(1/\delta)/q_k^2$ in \Cref{equ:concen}, it becomes:
    \begin{align*}
        \Pb\left(\hat{\mu}_k(\tau'_\delta)-\mu_k\geq \gamma_0\log^{-r'}(1/\delta) \right)\leq \frac{C_p}{\varsigma^p}.
    \end{align*}
    And for any $r'<1/8$, we can take $r_1=1/2+4r'<1$. Thus, $2r'<r_1/4$ leads to $\lim\limits_{\delta\rightarrow 0}\varsigma=\infty$ which concludes the proof by letting $\delta$ goes to zero.
\end{Proof}

In the following, we introduce the main proposition which shows the desired tracking result. 

\begin{proposition}\label{pro:deltaconverge}
    For $\alpha=1-\log^{-r}(1/\delta)$ with $0<r<1/2$, the sampling rule and stopping rule given in \Cref{for:stoppingrule2} with stopping time $\tau_\delta$ satisfies
    \begin{align*}
        \Pb\left(\lim\limits_{\delta\rightarrow 0}\frac{N_a(\tau_\delta)}{(1-\alpha)\tau_\delta}=u^*_a\right)=1,\quad a\in P_\calG(\bc,\bmu)
    \end{align*}
    and
    \begin{align*}
        \Pb\left(\lim\limits_{\delta\rightarrow 0}\frac{N_a(\tau_\delta)}{\tau_\delta}=\frac{1}{|N_\calG(\bc,\bmu)|}\right)=1,\quad a\in N_\calG(\bc,\bmu)
    \end{align*}
    and for arms out of $\supp(\calI(\bmu))$, 
    \begin{align*}
        \Pb\left(\lim\limits_{\delta\rightarrow 0}\frac{N_a(\tau_\delta)}{(1-\alpha)\tau_\delta}=0\right)=1,\quad a\notin \supp(\bmu).
    \end{align*}
    
\end{proposition}
\begin{Proofsketch}
    To prove this proposition, we focus on a special event $\Omega_0=\{\bar{\mu}_a(t)\mathop{\rightarrow}\limits_{t\rightarrow\infty}\mu_a, \forall a\}\cap \{\bar{c}_a(t)\mathop{\rightarrow}\limits_{t\rightarrow\infty}c_a, \forall a\}\cap \Omega_1\cap \Omega_2$ and $\Pb(\Omega_0)=1$. In this event, we can eliminate the effect of the force exploration during the sampling phase by Lemma~\ref{lem:cumulative-time}, i.e.
    \begin{align*}
        \max\limits_{1\leq a \leq K}\left|N_a(\tau_\delta)-\sum\limits_{s=0}^{\tau_\delta -1}\bu_{a}(\bc_\delta(s),\hat{\bmu}(s))\right|\leq K(1+\sqrt{\tau_\delta}).
    \end{align*}
    We can prove:
    \begin{align*}
        \lim\limits_{\delta\rightarrow 0} \frac{N_a(\tau_\delta)}{\tau_\delta(1-\alpha)}= \lim\limits_{\delta\rightarrow 0} \sum\limits_{s=0}^{\tau_\delta -1}\bu_{a}(\bc_\delta(s),\hat{\bmu}(s))/(\tau_\delta (1-\alpha)).
    \end{align*}
    We successfully change the original estimator $u_a^\epsilon$ to a more familiar estimator $u_a$ and the last is without force exploration.
    After that, we need to calculate 
    \begin{align*}
        \lim\limits_{\delta\rightarrow 0} \sum\limits_{s=0}^{\tau_\delta -1}\bu_{a}(\bc_\delta(s),\hat{\bmu}(s))/(\tau_\delta (1-\alpha)).
    \end{align*}
    Proposition~\ref{pro:bias-stop} shows a sub-linear commitment time of committing to $N(\bc)$, where the sub-linearity is with respect to the order of pulling times of arms in $P_\calG(\bc,\bmu)$. It means the total inaccurate time has a sub-linear order. With that, we introduce a result of sequential convergence in Lemma~\ref{lem:sequence} (the details of this lemma will be stated later):
    \begin{align*}
        S_\delta = b_1 +b _2+\cdots+b_{m_\delta}+f(\delta)(a_1+a_2+\cdots+a_{n_\delta-m_\delta})
        \xrightarrow[\lim\limits_{n\rightarrow\infty}a_n=a]{\text{some conditions}}
        \lim\limits_{\delta\rightarrow 0}S_\delta/(n_\delta f(\delta))=a
    \end{align*}
    and we can get the desired result.
\end{Proofsketch}
\begin{Proof}
    The critical point is proving the case of arm $a\in P_\calG(\bc,\bmu)$. First, we consider arm $a\in P_\calG(\bc,\bmu)$ and we need to prove
    \begin{align*}
        \Pb\left(\lim\limits_{\delta\rightarrow 0}\frac{N_a(\tau_\delta)}{(1-\alpha)\tau_\delta}=u^*_a\right)=1.
    \end{align*}
Denote event $\Omega_1=\{\lim\limits_{\delta\rightarrow 0}N(\bc_\delta(\tau'_\delta))=N(\bc)\}$ and from Proposition~\ref{pro:bias-stop}, $\Pb(\Omega_1)=1$.

By Lemma~\ref{lem:as-cumulativetime}, there exists a constant $K(\bmu)$ that the cumulative pulling time of arms in $P_\calG(\bc,\bmu)$ before algorithm stop satisfies
\begin{align*}
    \Pb\left(\lim\limits_{\delta\rightarrow 0}\frac{\sum_{a\in P_\calG(\bc,\bmu)}N_a(\tau_\delta)}{\log(1/\delta)}>K(\bmu)\right)=1,
\end{align*}
and, scaling with ratio $1-\alpha=\log^{r}(1/\delta)$, the total stopping time $\tau_\delta$ satisfies
\begin{align*}
    \Pb\left(\lim\limits_{\delta\rightarrow 0}\frac{\tau_\delta}{\log^{1+r}(1/\delta)}>K(\bmu)\right)=1,
\end{align*}
which we denoted as event $\Omega_2$.
Define event $\Omega_0=\{\bar{\mu}_a(t)\mathop{\rightarrow}\limits_{t\rightarrow\infty}\mu_a, \forall a\}\cap \{\bar{c}_a(t)\mathop{\rightarrow}\limits_{t\rightarrow\infty}c_a, \forall a\}\cap \Omega_1\cap \Omega_2$ and from the Law of Large Numbers and previous discussions, we have $\Pb(\Omega_0)=1$.

Recall that
\begin{align*}
    \boldsymbol{u}_\alpha (\bc_\delta(t),\hat{\bmu}(t))= \lv_\alpha(\bc_\delta(t),\bmu(t))+(1-\alpha)\boldsymbol{u}^*(\bc_\delta(t),\hat{\bmu}(t)).
\end{align*}
Using Lemma~\ref{lem:cumulative-time}, we have
\begin{align*}
    \max\limits_{1\leq a \leq K}\left|N_a(\tau_\delta)-\sum\limits_{s=0}^{\tau_\delta -1}\bu_{a}(\bc_\delta(s),\hat{\bmu}(s))\right|\leq K(1+\sqrt{\tau_\delta}).
\end{align*}
Thus,
\begin{align*}
    \max\limits_{1\leq a \leq K}\left|\frac{N_a(\tau_\delta)}{\tau_\delta(1-\alpha)}-\sum\limits_{s=0}^{\tau_\delta -1}\bu_{a}(\bc_\delta(s),\hat{\bmu}(s))/(\tau_\delta (1-\alpha))\right|\leq K\frac{1+\sqrt{\tau_\delta}}{\tau_\delta(1-\alpha)}.
\end{align*}
From the assumption of $\Omega_2$, $\tau_\delta=O(\log^{1+r}(1/\delta))$ and we have 
\begin{align*}
    \frac{1+\sqrt{\tau_\delta}}{\tau_\delta(1-\alpha)}\leq O\left(\frac{\log^{(1+r)/2}(1/\delta)}{\log(1/\delta)}\right)=O\left(\log^{(r-1)/2}(1/\delta)\right).
\end{align*}
Also $r<1$ leads to
\begin{align*}
    \lim\limits_{\delta\rightarrow 0}K\frac{1+\sqrt{\tau_\delta}}{\tau_\delta(1-\alpha)}=0.
\end{align*}
We get
\begin{align*}
    \lim\limits_{\delta\rightarrow 0} \frac{N_a(\tau_\delta)}{\tau_\delta(1-\alpha)}= \lim\limits_{\delta\rightarrow 0} \sum\limits_{s=0}^{\tau_\delta -1}\bu_{a}(\bc_\delta(s),\hat{\bmu}(s))/(\tau_\delta (1-\alpha)).
\end{align*}

For $a\in P_\calG(\bc,\bmu)$, denotes $\bu_a^*(\bc_\delta(s),\hat{\bmu}(s))=b_s$ be the $a$-th component of $\boldsymbol{u}^* (\bc_\delta(s),\hat{\bmu}(s))$. On event $\Omega_0$, we have $\lim\limits_{s\rightarrow\infty}b_s=u_a^*$ and $u_{\alpha, a}(\bc_\delta(s),\hat{\bmu}(s))=(1-\alpha)b_s$ where $u_{\alpha, a}(\bc_\delta(s),\hat{\bmu}(s))$ is $a$-th component of $\boldsymbol{u}_{\alpha} (\bc_\delta(s),\hat{\bmu}(s))$. Before time $\tau'_\delta$, the estimation of $N(\bc)$ and $P(\bc)$ might be inaccurate which will have influence on $u_{\alpha, a}(\bc_\delta(s),\hat{\bmu}(s))$ and we denote it as $b_s'$. Then, we have
\begin{align*}
    \lim\limits_{\delta\rightarrow 0}\frac{N_a(\tau_\delta)}{(1-\alpha)\tau_\delta}=u^*_a\Leftrightarrow \lim\limits_{\delta\rightarrow 0}\frac{b'_1+\cdots+b'_{\tau_\delta'}+(1-\alpha)(b_{\tau_\delta'+1}+\cdots+b_{\tau_\delta})}{\tau_\delta(1-\alpha)}=u_a^*
\end{align*}
and by $0<r<1$, we have
\begin{align*}
    \frac{\tau_\delta'}{\tau_\delta(1-\alpha)}=O(\log^{r_1-1}(1/\delta)),\quad \frac{\tau_\delta'}{\tau_\delta}=O(\log^{r_1-1-r}(1/\delta)).
\end{align*}
Thus, applying Lemma~\ref{lem:sequence}, we can get, for arm $a\in P_\calG(\bc,\bmu)$
\begin{align*}
    \Pb\left(\lim\limits_{\delta\rightarrow 0}\frac{N_a(\tau_\delta)}{(1-\alpha)\tau_\delta}=u^*_a\right)=1, \quad\forall a\in P_\calG(\bc,\bmu).
\end{align*}

For arms in $N_\calG(\bc,\bmu)$ and arms out of $\supp(\calI(\bmu))$ we can repeat the same process and we can conclude the proof. 
\end{Proof}

\begin{lemma}\label{lem:sequence}
    Let $\{a_n\}_{n\in \Nb}$, $\{b_n\}_{n\in \Nb}$ be two bounded sequences with $\lim_{n\rightarrow\infty}a_n=a$. For $\delta\in(0,1)$, $n_\delta,m_\delta$ are functions of $(0,1)\rightarrow\Nb$ with $n_\delta>m_\delta$ for each $\delta$ and $f(\delta)$ is a continuous function with $f(\delta)\neq 0$. Define $S_\delta$ as
    \begin{align*}
        S_\delta = b_1 +b _2+\cdots+b_{m_\delta}+f(\delta)(a_1+a_2+\cdots+a_{n_\delta-m_\delta}).
    \end{align*}
    When $m_\delta$ and $n_\delta$ satisfy $\lim_{\delta\rightarrow 0 }m_\delta/n_\delta = 0$ and $\lim_{\delta\rightarrow 0 }m_\delta/(f(\delta)n_\delta) = 0$, we have $\lim_{\delta\rightarrow 0}S_\delta/(n_\delta f(\delta))=a$.
\end{lemma}
\begin{Proof}
    Assume $|b_n|\leq B$ and we have
    \begin{align*}
        \frac{S_\delta}{n_\delta f(\delta)}=\frac{b_1+\cdots+b_{m_\delta}}{n_\delta f(\delta)}+\frac{a_1+\cdots+a_{n_\delta- m_\delta}}{n_\delta}.
    \end{align*}
    For the first term 
    \begin{align*}
        \frac{b_1+\cdots+b_{m_\delta}}{n_\delta f(\delta)}<B
        \cdot\frac{m_\delta}{n_\delta f(\delta)}\rightarrow 0,
    \end{align*}
   which indicates that
    \begin{align*}
        \lim\limits_{\delta\rightarrow 0}\frac{b_1+\cdots+b_{m_\delta}}{n_\delta f(\delta)}=0.
    \end{align*}
    Since $m_\delta/n_\delta\rightarrow 0$ and $m_\delta$ is always no less than $1$, we can get $n_\delta\rightarrow\infty$ and also
    \begin{align*}
        \lim\limits_{\delta\rightarrow 0}(n_\delta-m_\delta)=\lim\limits_{\delta\rightarrow 0}n_\delta\cdot(1-\frac{m_\delta}{n_\delta})=\infty.
    \end{align*}
    So, it holds that
    \begin{align*}
        \lim\limits_{\delta\rightarrow 0}\frac{a_1+\cdots+a_{n_\delta- m_\delta}}{n_\delta}&=\lim\limits_{\delta\rightarrow 0}\frac{a_1+\cdots+a_{n_\delta- m_\delta}}{n_\delta-m_\delta}\cdot\frac{n_\delta-m_\delta}{n_\delta}\\
        &=\lim\limits_{\delta\rightarrow 0}\frac{a_1+\cdots+a_{n_\delta- m_\delta}}{n_\delta-m_\delta}.
    \end{align*}
    Since $n_\delta-m_\delta\rightarrow\infty$, we only need to proof the following form
    \begin{align*}
        \lim\limits_{n\rightarrow\infty}\frac{a_1+a_2+\cdots+a_n}{n}=a.
    \end{align*}
    Assume $|a_n-a|<A, \forall n$, and for $\forall\varepsilon>0$, $\exists N_1$, s.t., $\forall n>N_1, |a_n-a|<\varepsilon/2$. When $n>N=\frac{2N_1A}{\varepsilon}$, we have 
    \begin{align*}
        \left|\frac{a_1+a_2+\cdots+a_n}{n}-a\right|&\leq \left|\frac{a_1-a}{n}+\cdots+\frac{a_n-a}{n}\right|\\
        &\leq \frac{|a_1-a|}{n}+\cdots+\frac{|a_n-a|}{n}\\
        &< \frac{N_1A}{n}+\frac{\varepsilon(n-N_1)}{2n}< \varepsilon,
    \end{align*}
    which concludes the proof of Lemma~\ref{lem:sequence}.
\end{Proof}

\section{PROOFS OF SECTION~\ref{sec:upperbound}}\label{app:section6}

\subsection{Proof of Proposition~\ref{thm:inf-thre}}
\begin{Proof}
    Denote $\varphi(\bmu)=m$ and $\calG_m=\cap_{(i,j)\in\calI_m}B_{ij}$. Let $T_{a,b}:=\inf\{t\in \mathbb{N}:Z_{a,b}(t)>\beta(t,\delta)\}$, one has
    \begin{align*}
        \Pb_{\bmu} (\tau_\delta < \infty, &\varphi(\hat{\bmu}(\tau_\delta))\neq \varphi(\boldsymbol{\mu}))\\
        &\leq \Pb_{\bmu}(\exists (a,b)\in \calI_m , \exists t\in\Nb:Z_{b,a}>\beta(t,\delta))\\
        &\leq \sum\limits_{(a,b)\in\calI_m}\Pb_\mu(T_{b,a}<\infty).
    \end{align*}
    As shown in \cite{garivier2016optimal}, we can have the following lemma.
    \begin{lemma}\label{lem:diff-ab}
        For any $a,b$ such that $\mu_a<\mu_b$, let $T_{a,b}:=\inf\{t\in \mathbb{N}:Z_{a,b}(t)>\beta(t,\delta)\}$ with $\beta(t,\delta)=\log(2tK(K-1)/\delta)$, we have
        \begin{align*}
            \Pb_\mu(T_{a,b}<\infty)\leq \frac{\delta}{K(K-1)}.
        \end{align*}
    \end{lemma}
    Thus, using Lemma~\ref{lem:diff-ab}, we have 
    \begin{align*}
       \Pb_\mu (\tau_\delta < \infty, \varphi(\hat{\bmu}(\tau_\delta))\neq\varphi(\bmu))\leq |\calI_m|\cdot\frac{\delta}{K(K-1)}\leq\delta.
    \end{align*}
    which completes the proof.
\end{Proof}

\subsection{Proof of Proposition~\ref{pro:Dev-Thre}}
\begin{Proof}
    It can be verified that if $x>y$,
    \begin{align*}
        I_\alpha(x,y)=\inf\limits_{x'<y'}\big[\alpha d(x,x')+(1-\alpha)d(y,y')\big].
    \end{align*}
    For every $a,b$ that $\mu_a<\mu_b$ and $\hat{\mu}_a(t)>\hat{\mu}_a(t)$, we have
    \begin{align*}
        Z_{a,b}(t)&=(N_a(t)+N_b(t))I_{\frac{N_a(t)}{N_a(t)+N_b(t)}}(\hat{\mu}_a(t),\hat{\mu}_b(t))\\
        &=\inf\limits_{\mu_a'<\mu_b'}N_a(t)d(\hat{\mu}_a(t),\mu_a')+N_b(t)d(\hat{\mu}_b(t),\mu_b')\\
        &\leq N_a(t)d(\hat{\mu}_a(t),\mu_a)+N_b(t)d(\hat{\mu}_b(t),\mu_b).
    \end{align*}
    For $\mu_a>\mu_b$, repeating the previous discussion, we have the same formula:
    \begin{align*}
        Z_{b,a}(t)\leq N_a(t)d(\hat{\mu}_a(t),\mu_a)+N_b(t)d(\hat{\mu}_b(t),\mu_b).
    \end{align*}
    Assuming $\boldsymbol{\mu}=(\mu_1,\mu_2,\ldots,\mu_K)$ and $\varphi(\bmu)=m$, $\calG_m=\cap_{(i,j)\in\calI_m}B_{i,j}$, we have
    \begin{align*}
        &\Pb_{\boldsymbol{\mu}}(\tau_\delta\leq \infty , \varphi(\hat{\bmu}(\tau_\delta))\neq \varphi(\boldsymbol{\mu}))\\
        \leq &\Pb_{\boldsymbol{\mu}}(\exists (a,b)\in \calI_m , \exists t\in\Nb:\hat{\mu}_b(t)>\hat{\mu}_a(t),Z_{b,a}>\beta(t,\delta))\\
        \leq &\Pb_{\boldsymbol{\mu}}(\exists (a,b)\in \calI_m, \exists t\in \mathbb{N} : N_a(t)d(\hat{\mu}_a(t),\mu_a)+N_b(t)d(\hat{\mu}_b(t),\mu_b)\geq\beta(t,\delta))\\
        \leq &\Pb_{\boldsymbol{\mu}}\left(\exists t\in \mathbb{N} : \sum\limits_{a=1}^K N_a(t)d(\hat{\mu}_a(t),\mu_a)\geq\beta(t,\delta)\right)\\
        \leq &\sum\limits_{t=1}^\infty e^{K+1}\left(\frac{\beta(t,\delta)^2\log (t)}{K}\right)^Ke^{-\beta(t,\delta)}.
    \end{align*}
    The last inequality follows from \cite{Magureanu2014} whose result can straightforwardly generalize to a one-parameter exponential family. With a rate form $\beta(t,\delta)=\log(Ct^\theta/\delta)$, we need to prove that for any $\delta \in (0, 1)$, there exists $C = O(\text{polylog}(1/\delta))$ in $\beta(t, \delta) = \log(C t^{\theta} / \delta)$ such that 
    \begin{align*}
        \sum\limits_{t=1}^\infty e^{K+1}\left(\frac{\beta(t,\delta)^2\log (t)}{K}\right)^Ke^{-\beta(t,\delta)}\leq \delta.
    \end{align*}
    It is equivalent to prove that for any $\delta \in (0, 1)$ there exists a function of $\delta$, $C = C(\delta) = O(\text{polylog}(1/\delta))$, such that 
    \begin{align*}
        \sup_{\delta \in (0, 1)} \sum\limits_{t=1}^\infty \frac{e^{K+1}}{K^K}\frac{(\log^2(C t^\theta/\delta)\log t)^K}{C t^\theta} \leq 1.
    \end{align*}
    We can prove this by taking $C = C_0(e +\log^{2K + 1}(1/\delta))=C_0 f_0(\delta)$ with a sufficiently large constant $C_0$. 

    For any $t\in \mathbb{N}, \delta\in(0,1)$, we have 
    \begin{align*}
        f_0(\delta) t^\theta/\delta> e.
    \end{align*}
    Take sufficiently large $C_0$ that 
    \begin{align*}
        \log(C_0)\geq 1+\frac{1}{\log(f_0(\delta) t^\theta/\delta)-1},
    \end{align*}
    which implies 
    \begin{align}
        \log(C_0f_0(\delta) t^\theta/\delta) = \log(C_0)+\log(f_0(\delta) t^\theta/\delta) \leq \log(C_0) \log(f_0(\delta) t^\theta/\delta). 
    \end{align}
    Similarly, we have
    \begin{align}
        \log(f_0(\delta) t^\theta/\delta) \leq \log(f_0(\delta)/\delta) \log( t^\theta). 
    \end{align}
    It then follows that 
    \begin{align*}
        & \sum\limits_{t=1}^\infty \frac{e^{K+1}}{K^K} \frac{(\log^2( C_0f_0(\delta) t^\theta/\delta)\log t)^K}{ C_0f_0(\delta) t^\theta} \\
        & \leq \frac{e^{K+1}}{K^K} \sum\limits_{t=1}^\infty \frac{(\log^2(C_0)\cdot\log^2(f_0(\delta)
        t^\theta/\delta)\log t)^K}{ C_0 f_0(\delta) t^\theta} \\
        & \leq \frac{\log^{2K}(C_0)}{C_0} \frac{e^{K+1}}{K^K} \frac{\log^{2K}(f_0(\delta)/\delta)}{f_0(\delta)}\sum\limits_{t=1}^\infty \frac{\log^{2K}( t^\theta)\log^K(t)}{t^\theta}. 
    \end{align*}
    Note that the series $\sum\limits_{t=1}^\infty \frac{\log^{2K}( t^\theta)\log^K(t)}{t^\theta}$ is bounded, and the function $\frac{\log^{2K}(f_0(\delta)/\delta)}{f_0(\delta)}$ is also uniformly bounded since it is continuous and takes $0$ and $1/e$ when $\delta$ approaching to $0$ and $1$, respectively. We can then establish the proposition by taking a sufficiently large $C_0$ to guarantee that 
    \begin{align*}
        \frac{\log^{2K}(C_0)}{C_0} \frac{e^{K+1}}{K^K} \frac{\log^{2K}(f_0(\delta)/\delta)}{f_0(\delta)}\sum\limits_{t=1}^\infty \frac{\log^{2K}( t^\theta)\log^K(t)}{t^\theta} \leq 1.
    \end{align*}
\end{Proof}

\subsection{Proof of Theorem~\ref{thm:upperbound}}
By Proposition~\ref{pro:deltaconverge}, those arms not in $\supp(\calI)$ are pulled with {sub-linear times} with respect to arms in $P_\calG(\bc,\bmu)$) which will only influence $o(\log(1/\delta))$ cumulative cost and will not affect the result. Thus, in the proof of the upper bound theorem, we can assume $\supp(\calI(\bmu))=\calA$. So, $P_\calG(\bc,\bmu)=P(\bc)$, $N_\calG(\bc,\bmu)=N(\bc)$ and $\Eb^\calG[f(\bc,\bmu;\tau_\delta)]=\Eb[f(\bc,\bmu;\tau_\delta)]$

We first introduce the concept of the sample distribution function, we say 
\begin{align}\label{for:distribution}
    g:\mathbb{R}^K\times\mathbb{R}^K\rightarrow \Sigma_K
\end{align}
is a sample distribution, if $g(\hat{c}(t),\hat{\mu}(t))$ converge to some value $g^*$ in $\Sigma_K$ when $t\rightarrow \infty$.

For any sample distribution function $g:\mathbb{R}^K\times\mathbb{R}^K\rightarrow \Sigma_K$, $g(\hat{c}(t), \hat{\mu}(t))$ calculates the desired ratio of pulling arms, and the pulling strategy will be determined by the desired pulling ratio. As $t \rightarrow \infty$, we have $\hat{c}(t) \rightarrow c$, $\hat{\mu} \rightarrow \mu$ and $g(\hat{c}(t), \hat{\mu}(t))$ converges to the desired pulling ratios $g^*$. 

Specifically, the sampling rule under the specific sample distribution function is as follows: we still calculate the $L^\infty$ projection of $g$ denoted by $g^\epsilon $. The arm pulled at time step $t+1$ is $A_{t+1}\in \argmax\limits_{1\leq a\leq K}\sum\limits_{s=0}^t g_a^{\epsilon_s}(\hat{c}(s),\hat{\mu}(s))-N_a(t)$. The projection allows the sampling rule to have enough exploration power by $\epsilon$. In this case, $\sum\limits_{s=0}^t g_a^{\epsilon_s}(\hat{c}(s),\hat{\mu}(s))$ is the desired number of pulls of arm $a$ and $N_a(t)$ is the actual number of pulls of arm $a$. The sampling strategy
\begin{align*}
    A_{t+1}\in \argmax\limits_{1\leq a\leq K}\sum\limits_{s=0}^t g_a^{\epsilon_s}(\hat{c}(s),\hat{\mu}(s))-N_a(t)
\end{align*}
means to pull the arm whose amount of pulls is the most lagging behind the desired number of pulls. 

Before proving the main theorem, we need to have some preparations. 
First, we consider using the sampling rule under the sample distribution function $g$ discussed in \Cref{for:distribution}.
As for the lower bound, repeat the discussion in Section~\ref{sec:sample}, if $g$ converges to $\boldsymbol{\omega}\in \Sigma_K$ we have
\begin{align*}
    \mathbb{E}_{\bmu}[\tau_\delta]\geq T^*(\boldsymbol{\omega},\boldsymbol{\mu})\cdot \kl(\delta,1-\delta)
\end{align*}
where
\begin{align*}
    T^*(\boldsymbol{\omega},\boldsymbol{\mu})^{-1}=\inf\limits_{\boldsymbol{\lambda}\in \Alt(\boldsymbol{\mu})}\left(\sum\limits_{a=1}^K\omega_ad(\mu_a,\lambda_a)\right)=\min\limits_{(a,b)\in\calI}(\omega_b+\omega_a)I_{\frac{\omega_b}{\omega_b+\omega_a}}(\mu_b,\mu_a).
\end{align*}
\begin{remark}\label{rem:alt}
    We say $\Alt(\boldsymbol{\mu})$ with out $\boldsymbol{c}$ means that $\Alt(\boldsymbol{\mu})=\{\blambda\in S:\ \varphi(\blambda)\neq\varphi(\bmu)\}$.
\end{remark}

\begin{theorem}[upper bound for sample complexity]\label{thm:oldupbound}
     Let $\boldsymbol{\mu}$ be an exponential family bandit model. Let $\theta \in [1,e/2]$ and $r(t)=O(t^\theta)$. Using stopping rule given in \Cref{for:stoppingrule2} with $\beta(t,\delta)=\log(r(t)/\delta)$, and the sampling rule under any sample distribution function $g$, we have
    \begin{align*}
        \limsup\limits_{\delta \rightarrow 0}\dfrac{\mathbb{E}_{\boldsymbol{\mu}}[\tau_\delta]}{\log(1/\delta)}\leq \theta T^*(g^*,\bmu).
    \end{align*}
\end{theorem}
This theorem is proved in Theorem 14 of \cite{garivier2016optimal}, with some notation changes 
and model assumption change as $\xi=\xi(\epsilon)< \min\limits_{(a,b)\in\calI}|\mu_a-\mu_b|/2$, $\mathcal{I}_\epsilon :=[\mu_1-\xi,\mu_1+\xi]\times\cdots \times[\mu_K-\xi,\mu_K+\xi]$,  
and the stopping statistic in the stopping rule given in \Cref{for:stoppingrule2} changes to 
\begin{align*}
    \min\limits_{(a,b)\in\calI}|Z_{a,b}(t)|&=\min[\min\limits_{\mu_a<\mu_b\atop (a,b)\in\calI}Z_{b,a},\min\limits_{\mu_a>\mu_b\atop (a,b)\in\calI}Z_{a,b}(t)]\\
    &=\min\Big[\min\limits_{\mu_a<\mu_b\atop (a,b)\in\calI}\big[N_b(t)d(\hat{\mu}_b(t),\hat{\mu}_{b,a}(t))+N_a(t)d(\hat{\mu}_a(t),\hat{\mu}_{b,a}(t))\big],\\
    &\ \ \ \ \ \ \ \ \ \ \ \min\limits_{\mu_a>\mu_b\atop (a,b)\in\calI}\big[N_a(t)d(\hat{\mu}_a(t),\hat{\mu}_{a,b}(t))+N_b(t)d(\hat{\mu}_b(t),\hat{\mu}_{a,b}(t))\big]\Big].
\end{align*}
Also, the function is rewritten as 
\begin{align*}
    \hat{g}(\boldsymbol{\mu}',\boldsymbol{\omega}')=\min\limits_{(a,b)\in \calI}(\omega_b'+\omega_a')I_{\frac{\omega_b'}{\omega_b'+\omega_a'}}(\mu_b',\mu_a')
\end{align*}
using $(\omega_b'+\omega_a')I_{\frac{\omega_b'}{\omega_b'+\omega_a'}}(\mu_b',\mu_a')=(\omega_a'+\omega_b')I_{\frac{\omega_a'}{\omega_a'+\omega_b'}}(\mu_a',\mu_b')$.

Now, we give the proof of the Lemma~\ref{lem:ratio} of the main paper:
\begin{align}\label{for:ratio2}
    \lim\limits_{\delta\rightarrow 0}T^*(\boldsymbol{u}_\alpha^*,\boldsymbol{\mu})^{-1}\dfrac{\mathbb{E}_\mu[\tau_\delta]}{\mathbb{E}[f(\boldsymbol{c}, \boldsymbol{\mu};\tau_\delta)]}=T^*(\boldsymbol{c},\boldsymbol{\mu})^{-1}.
\end{align}
Since those arms not in $\supp(\calI(\bmu))$ incur only sub-linear regret (with respect to arms in $P_\calG(\bc,\bmu)$), which will only influence $o(\log(1/\delta))$ of the cumulative cost, we can assume $\supp(\calI(\bmu))=\calA$ in this proof. 
\begin{Proof}
    Recall $\boldsymbol{u}_\alpha^*=(u_1,\ldots,u_K)$ and $\boldsymbol{u}^*(\boldsymbol{c},\boldsymbol{\mu})=(u^*_1,\ldots,u^*_K)$. 
    When $\delta\rightarrow 0$, the pulling time $N_a(\tau_\delta)\rightarrow \infty$, so the sample distribution converges to $\boldsymbol{u}_\alpha$. Thus, we get
    \begin{align*}
        \lim\limits_{\delta\rightarrow 0}\frac{\mathbb{E}[f(\boldsymbol{c}, \boldsymbol{\mu};\tau_\delta)]}{\mathbb{E}_\mu[\tau_\delta]}=\frac{\sum_{i\in P(\boldsymbol{c})}c_i\Eb[N_i(\tau_\delta)]}{\sum_{i=1}^K\Eb[N_i(\tau_\delta)]}=(1-\alpha)(\sum\limits_{i\in P(\boldsymbol{c})}u^*_ic_i).
    \end{align*}
    Also, when $\delta\rightarrow 0$, $\alpha$ will converge to $1$. For each $t\in P(\boldsymbol{c}),\ s\in N(\boldsymbol{c})$, by the continuity of $d(x,y)$, we have
    \begin{align*}
        &\lim\limits_{\alpha \rightarrow 1}\frac{\alpha/|N(\boldsymbol{c})|+u_t}{1-\alpha}I_{\frac{\alpha/|N(\boldsymbol{c})|}{\alpha/|N(\boldsymbol{c})|+u_t}}(\mu_s,\mu_t)=\lim\limits_{\alpha\rightarrow 1}\frac{1}{(1-a)|N(\boldsymbol{c})|}I_{\frac{\alpha}{\alpha+u_t|N(\boldsymbol{c})|}}(\mu_s,\mu_t)\\
        =&\lim\limits_{\alpha\rightarrow 1}\bigg(\frac{1}{(1-\alpha)|N(\boldsymbol{c})|}d(\mu_s,\frac{\alpha}{\alpha+u_t|N(\boldsymbol{c})|}\mu_s+\frac{u_t|N(\boldsymbol{c})|}{\alpha+u_t|N(\boldsymbol{c})|}\mu_t)+\\
        &\quad\quad\frac{1}{(1-\alpha)|N(\boldsymbol{c})|}\frac{u_t|N(\boldsymbol{c})|}{\alpha+u_t|N(\boldsymbol{c})|}d(\mu_t,\frac{\alpha}{\alpha+u_t|N(\boldsymbol{c})|}\mu_s+\frac{u_t|N(\boldsymbol{c})|}{\alpha+u_t|N(\boldsymbol{c})|}\mu_t)\bigg)\\
        =&\lim\limits_{\alpha\rightarrow 1}\frac{1}{(1-\alpha)|N(\boldsymbol{c})|}d(\mu_s,\frac{\alpha}{\alpha+u_t|N(\boldsymbol{c})|}\mu_s+\frac{u_t|N(\boldsymbol{c})|}{\alpha+u_t|N(\boldsymbol{c})|}\mu_t)+\frac{u_t}{(1-\alpha)}d(\mu_t,\mu_s)\\
        =&\lim\limits_{\alpha\rightarrow 1}\frac{1}{(1-\alpha)|N(\boldsymbol{c})|}d(\mu_s,\frac{\alpha}{\alpha+u_t|N(\boldsymbol{c})|}\mu_s+\frac{u_t|N(\boldsymbol{c})|}{\alpha+u_t|N(\boldsymbol{c})|}\mu_t)+u^*_td(\mu_t,\mu_s)
    \end{align*}
    
    Using L'Hopital rule and $\frac{d}{dy}d(x,y)=(y-x)/b''(b^{-1}(y))$, the first term becomes
    \begin{align*}
        \lim\limits_{\alpha\rightarrow 1}-\frac{(\mu_t-\mu_s)^2u_t^2|N(\boldsymbol{c})|^2}{(a+u_tN(\boldsymbol{c}))^3}\cdot \frac{1}{b''(b^{-1}(y))}/(-1)=0.
    \end{align*}
    Since $u_t=(1-\alpha)u^*_t\rightarrow 0$, we have
    {
    \small
    \begin{align*}
        \lim\limits_{\alpha \rightarrow 1}\frac{\alpha/|N(\boldsymbol{c})|+u_t}{1-\alpha}I_{\frac{\alpha/|N(\boldsymbol{c})|}{\alpha/|N(\boldsymbol{c})|+u_t}}(\mu_s,\mu_t)=u^*_td(\mu_t,\mu_s). 
    \end{align*}
    }
    Respectively, we have
    {
    \small
    \begin{align*}
        \lim\limits_{\alpha \rightarrow 1}\frac{u_s+\alpha/|N(\boldsymbol{c})|}{1-\alpha}I_{\frac{u_s}{u_s+\alpha/|N(\boldsymbol{c})|}}(\mu_s,\mu_t)=u^*_sd(\mu_s,\mu_t). 
    \end{align*}
    }
    Now, we can compute the concerning formula as
    \begin{align*}
        &\lim\limits_{\delta\rightarrow 0}T^*(\boldsymbol{u}_\alpha^*,\boldsymbol{\mu})^{-1}\frac{\mathbb{E}_\mu[\tau_\delta]}{\mathbb{E}[f(\boldsymbol{c}, \boldsymbol{\mu};\tau_\delta)]}\\=&T^*(\boldsymbol{u}_\alpha^*,\boldsymbol{\mu})^{-1}\bigg[(1-\alpha)(\sum\limits_{i\in P(\boldsymbol{c})}u^*_i c_i)\bigg]^{-1}\\
        =&\lim\limits_{\alpha\rightarrow 1}\min\big[\ T_{11}(\boldsymbol{c},\boldsymbol{u}_\alpha^*)^{-1},T_{12}(\boldsymbol{c},\boldsymbol{u}_\alpha^*)^{-1},T_{21}(\boldsymbol{c},\boldsymbol{u}_\alpha^*)^{-1} \big]\cdot \bigg[(1-\alpha)(\sum\limits_{i\in P(\boldsymbol{c})}u^*_i c_i)\bigg]^{-1}
     \end{align*}
    where
    \begin{align*}
        T_{11}(\boldsymbol{c},\boldsymbol{u}_\alpha^*)^{-1}&=\min\limits_{(a,b)\in\calI\atop a,b\notin N(\boldsymbol{c})}\ (u_b+u_a)I_{\frac{u_b}{u_b+u_a}}(\mu_b,\mu_a),\\
        T_{12}(\boldsymbol{c},\boldsymbol{u}_\alpha^*)^{-1}&=\min\limits_{(a,b)\in\calI\atop a\notin N(\boldsymbol{c}),\ b\in N(\boldsymbol{c})}(\frac{\alpha}{|N(\boldsymbol{c})|}+u_a)I_{\frac{\alpha/|N(\boldsymbol{c})|}{\alpha/|N(\boldsymbol{c})|+u_a}}(\mu_b,\mu_a),\\
        T_{21}(\boldsymbol{c},\boldsymbol{u}_\alpha^*)^{-1}&=\min\limits_{(a,b)\in\calI\atop b\notin N(\boldsymbol{c}),\ a\in N(\boldsymbol{c})}(u_b+\frac{\alpha}{|N(\boldsymbol{c})|})I_{\frac{u_b}{u_b+\alpha/|N(\boldsymbol{c})|}}(\mu_b,\mu_a).
    \end{align*}
    From the previous discussion, we have 
    \begin{align*}
        \lim\limits_{\alpha\rightarrow 1}T_{11}(\boldsymbol{c},\boldsymbol{u}_\alpha^*)^{-1}&\cdot \bigg[(1-\alpha)\bigg(\sum\limits_{i\in P(\boldsymbol{c})}u^*_i c_i\bigg)\bigg]^{-1}\\
        =&\min\limits_{(a,b)\in\calI\atop a,b\notin N(\boldsymbol{c})}\ (u^*_b+u^*_a)I_{\frac{u^*_b}{u^*_b+u^*_a}}(\mu_b,\mu_a)\bigg(\sum\limits_{i\in P(\boldsymbol{c})}u^*_i c_i\bigg)^{-1}\\
        \lim\limits_{\alpha\rightarrow 1}T_{12}(\boldsymbol{c},\boldsymbol{u}_\alpha^*)^{-1}&\cdot \bigg[(1-\alpha)\bigg(\sum\limits_{i\in P(\boldsymbol{c})}u^*_i c_i\bigg)\bigg]^{-1}\\
        =&\lim\limits_{\alpha\rightarrow 1}\min\limits_{(a,b)\in\calI\atop a\notin N(\boldsymbol{c}),\ b\in N(\boldsymbol{c})}\frac{\alpha/|N(\boldsymbol{c})|+u_a}{1-\alpha}I_{\frac{\alpha/|N(\boldsymbol{c})|}{\alpha/|N(\boldsymbol{c})|+u_a}}(\mu_b,\mu_a)\bigg(\sum\limits_{i\in P(\boldsymbol{c})}u^*_i c_i\bigg)^{-1}\\
        =&\min\limits_{(a,b)\in\calI\atop a\notin N(\boldsymbol{c}),\ b\in N(\boldsymbol{c})}u^*_ad(\mu_a,\mu_b)\bigg(\sum\limits_{i\in P(\boldsymbol{c})}u^*_i c_i\bigg)^{-1}\\
        \lim\limits_{\alpha\rightarrow 1}T_{21}(\boldsymbol{c},\boldsymbol{u}_\alpha^*)^{-1}&\cdot \bigg[(1-\alpha)\bigg(\sum\limits_{i\in P(\boldsymbol{c})}u^*_i c_i\bigg)\bigg]^{-1}\\
        =&\lim\limits_{\alpha\rightarrow 1}\min\limits_{(a,b)\in\calI\atop b\notin N(\boldsymbol{c}),\ a\in N(\boldsymbol{c})}\ \frac{u_b+\alpha/|N(\boldsymbol{c})|}{1-\alpha}I_{\frac{u_b}{u_b+\alpha/|N(\boldsymbol{c})|}}(\mu_b,\mu_a)\bigg(\sum\limits_{i\in P(\boldsymbol{c})}u^*_i c_i\bigg)^{-1}\\
        =&\min\limits_{(a,b)\in\calI\atop b\notin N(\boldsymbol{c}),\ a\in N(\boldsymbol{c})}u^*_bd(\mu_b,\mu_a)\bigg(\sum\limits_{i\in P(\boldsymbol{c})}u^*_i c_i\bigg)^{-1}
    \end{align*}
    and combining
    \begin{align*}
        (\frac{\omega^*_b}{c_b}+\frac{\omega^*_a}{c_a})\bigg(\sum u^*_ic_i\bigg)&=(\frac{\omega^*_b}{c_b}+\frac{\omega^*_a}{c_a})\cdot \frac{1}{\sum_{i\in P(\boldsymbol{c})}\frac{\omega^*_i}{c_i}}=u^*_a+u^*_b,\ a,b\in P(\boldsymbol{c})\\
        \frac{\omega^*_a}{c_a}\bigg(\sum u^*_ic_i\bigg)&=\frac{\omega^*_a}{c_a}\cdot \frac{1}{\sum_{i\in P(\boldsymbol{c})}\frac{\omega^*_i}{c_i}}=u^*_a,\ a\in P(\boldsymbol{c})
    \end{align*}
    we get
    \begin{align*}
        \lim\limits_{\delta\rightarrow 0}T^*(\boldsymbol{u}_\alpha^*,\boldsymbol{\mu})^{-1}\bigg[(1-\alpha)\bigg(\sum\limits_{i\in P(\boldsymbol{c})}u^*_i c_i\bigg)\bigg]^{-1}=T^*(\boldsymbol{c},\boldsymbol{\mu})^{-1}
    \end{align*}
    which can be written as
    \begin{align*}
        \lim\limits_{\delta\rightarrow 0}T^*(\boldsymbol{u}_\alpha^*,\boldsymbol{\mu})^{-1}\frac{\mathbb{E}_\mu[\tau_\delta]}{\mathbb{E}[f(\boldsymbol{c}, \boldsymbol{\mu};\tau_\delta)]}=T^*(\boldsymbol{c},\boldsymbol{\mu})^{-1}.
    \end{align*}
\end{Proof}

After proving this lemma, combining Theorem~\ref{thm:oldupbound}, we can finish the proof of the upper bound theorem (which is Theorem~\ref{thm:upperbound} in the main paper).
\begin{Proof}
We directly compute the result:
\begin{align*}
    \limsup\limits_{\delta \rightarrow 0}\dfrac{\mathbb{E}[f(\boldsymbol{c}, \boldsymbol{\mu};\tau_\delta)]}{\log(1/\delta)}&=\limsup\limits_{\delta \rightarrow 0}\dfrac{\mathbb{E}_\mu[\tau_\delta]}{\log(1/\delta)}\cdot \dfrac{\mathbb{E}[f(\boldsymbol{c}, \boldsymbol{\mu};\tau_\delta)]}{\mathbb{E}_\mu[\tau_\delta]}\\
    &\leq \lim\limits_{\delta\rightarrow 0}\theta T^*(\boldsymbol{u}_\alpha^*,\boldsymbol{\mu})\dfrac{\mathbb{E}[f(\boldsymbol{c}, \boldsymbol{\mu};\tau_\delta)]}{\mathbb{E}_\mu[\tau_\delta]}\\
    &= \theta T^*(\boldsymbol{c},\boldsymbol{\mu})
\end{align*}
which concludes the proof.
\end{Proof}

\section{ALMOST-SURE UPPER BOUND}
In this section, we derive the almost-sure upper bound which, as Theorem~\ref{thm:upperbound} indicates, also characterizes the properties of cumulative cost.

\begin{theorem}[Almost Sure Upper Bound]\label{thm:as-upper} 
    Let $\theta\in [1,e/2]$ and $r(t)=O(t^\theta)$. Using stopping rule given in \Cref{for:stoppingrule2} with $\beta(t,\delta)=\log(r(t)/\delta)$, the CAET algorithm ensures
    \begin{align*}
        \Pb_\mu\left(\limsup\limits_{\delta\rightarrow 0}\frac{f(\boldsymbol{c},\boldsymbol{\mu};\tau_\delta)}{\log(1/\delta)}\leq \theta T^*(\boldsymbol{c},\boldsymbol{\mu})\right)=1.
    \end{align*}
\end{theorem}
\begin{Proof}
    For simplicity, we still assume $\supp(\calI(\bmu))=\calA$ with the reason stated as before. When $t$ goes to infinity, the actual sampling distribution is 
    \begin{align*}
    \boldsymbol{u}_\alpha^*=\boldsymbol{u}_{\alpha(\delta)}^*=1_{\alpha(\delta)}(\boldsymbol{c})+(1-\alpha(\delta))\boldsymbol{u}^*(\boldsymbol{c},\boldsymbol{\mu}):=(u_1,u_2,\ldots,u_K).
    \end{align*}
    Define $\calE$ as
    \begin{align*}
        \calE=\{\forall a\in \calA, N_a(t)/t\mathop{\rightarrow}\limits_{t\rightarrow\infty}u_a\}\cup\{\hat{\boldsymbol{\mu}}(t)\mathop{\rightarrow}\limits_{t\rightarrow\infty}\boldsymbol{\mu}\}\cup\{\hat{\boldsymbol{c}}(t)\mathop{\rightarrow}\limits_{t\rightarrow\infty}\boldsymbol{c}\}.
    \end{align*}
    From the Proposition~\ref{pro:time-converge} and the Law of Large Numbers, event $\calE$ happens with probability 1. We have:
    \begin{align*}
        \frac{f(\boldsymbol{c},\boldsymbol{\mu};\tau_\delta)}{\tau_\delta} = \sum\limits_{a\in P(\bc)} c_a \frac{N_a(\tau_\delta)}{\tau_\delta}.
    \end{align*}
    And, in event $\calE$, $N_a(t)/t\mathop{\rightarrow}\limits_{t\rightarrow\infty}u_a$ and $\tau_\delta$ will goes to $\infty$ as the lower bound of $\tau_\delta$ implies. So,
    \begin{align}\label{for:almostcon}        \Pb\left(\frac{f(\boldsymbol{c},\boldsymbol{\mu};\tau_\delta)}{\tau_\delta}\xrightarrow[\delta\rightarrow 0]{} (1-\alpha)\sum_{a\in P(\bc)}c_au_a^*\right)=1.
    \end{align}
    Define
    \begin{align*}
        T^*(\boldsymbol{u}_\alpha^*,\boldsymbol{\mu})^{-1}=\min\limits_{(a,b)\in \calI}\inf\limits_{\lambda\in S:\atop(\lambda_a-\lambda_b)(\mu_a-\mu_b)<0}\sum\limits_{a=1}^Ku_ad(\mu_a,\lambda_a)    =\min\limits_{(a,b)\in\calI}\ (u_a+u_b)I_{\frac{u_b}{u_b+u_a}}(\mu_b,\mu_a).
    \end{align*}
    According to Proposition 13 in \cite{garivier2016optimal}, we have $\tau_\delta$ is bounded for any $\delta > 0$ and 
    \begin{align*}
        \limsup\limits_{\delta\rightarrow 0}\frac{\tau_\delta}{\log(1/\delta)}\leq (1+\epsilon)\theta T^*(\boldsymbol{u}_\alpha^*,\boldsymbol{\mu}).
    \end{align*}
    Also, Lemma~\ref{lem:ratio} in the main paper, which is \Cref{for:ratio2} in this Appendix, tells us that 
    \begin{align*}
        \lim\limits_{\delta\rightarrow 0}T^*(\boldsymbol{u}_\alpha^*,\boldsymbol{\mu})^{-1}(1-\alpha)\sum_{a\in P(\bc)}c_au_a^*=T^*(\boldsymbol{c},\boldsymbol{\mu})^{-1}.
    \end{align*}
    Combining with Equation~\eqref{for:almostcon}, we get
    \begin{align*}
        \limsup\limits_{\delta\rightarrow 0}\frac{f(\boldsymbol{c},\boldsymbol{\mu};\tau_\delta)}{\log(1/\delta)}&=\limsup\limits_{\delta\rightarrow 0}\frac{\tau_\delta}{\log(1/\delta)}\cdot \frac{f(\boldsymbol{c},\boldsymbol{\mu};\tau_\delta)}{\tau_\delta}\\
        &\leq (1+\epsilon)\theta\limsup\limits_{\delta\rightarrow 0}T^*(\boldsymbol{u}_\alpha^*,\boldsymbol{\mu})\cdot\frac{f(\boldsymbol{c},\boldsymbol{\mu};\tau_\delta)}{\tau_\delta}\\
        &=(1+\epsilon)\theta T^*(\boldsymbol{c},\boldsymbol{\mu}).
    \end{align*}
    Letting $\epsilon$ go to zero concludes the proof.
\end{Proof}

\section{SPECIAL CASES}
\label{sec:ranking}

In this section, we leverage the theory developed above to study three important special cases of interest -- best arm identification, ranking identification, and regret minimization, by specifying the parameters (identification task $\mathcal{I}$ and cost vector $\bc$) in the general scenario.

The cost-aware scenario of BAI was studied in a concurrent work \citep{kanarios2024cost}, yet for a strictly positive cost vector. In contrast, we consider more general identification problems with non-negative cost vectors. Note that allowing certain costs to be zero leads to more technical challenges and allows the generality of the scenarios. For example, the regret minimization case studied in Section~\ref{sec:regretmin} of the main paper, can be viewed as a case with a non-negative cost vector.

\subsection{Best Arm Identification}

In the best arm identification (BAI) problem, the goal is to identify the arm with the largest expected reward. In the pure-exploration problem $(\mathcal{G}, \varphi)$ of identifying the best arm, $M = K$ and $\mathcal{G}_k = \cap_{j \not= k} \mathcal{B}_{kj}$, which is a pairwise exploration task. With a positive cost vector $\bc=(c_1,\ldots,c_K)$, our theorem implies, for any $\delta$-PAC algorithm
\begin{align*}
    \Eb[f(\bc,\bmu;\tau_\delta)]\geq T^*(\bc,\bmu)\kl(\delta,1-\delta)
\end{align*}
where
\begin{align*}
    T^*(\bc,\bmu)^{-1}=\sup\limits_{\boldsymbol{\omega}\in\Sigma_K}\inf\limits_{\blambda\in \Alt(\bmu)}\sum_{a\in\calA}\frac{\omega_a}{c_a}d(\mu_a,\lambda_a).
\end{align*}
Here $c_i>0$, so actually $\Alt(\bc,\bmu)=\Alt(\bmu)$. And our CAET algorithm satisfies
\begin{align*}
    \limsup\limits_{\delta \rightarrow 0}\frac{\mathbb{E}[f(\boldsymbol{c}, \boldsymbol{\mu};\tau_\delta)]}{\log(1/\delta)}\leq \theta T^*(\boldsymbol{c},\boldsymbol{\mu})
\end{align*}
recovering the results in \cite{kanarios2024cost}. Also, when we take $\bc=(1,\ldots,1)$, the cost $\Eb[\bc,\bmu;\tau_\delta]$ become sample time $\Eb_{\bmu}[\tau_\delta]$ and gets
\begin{align*}
    \limsup\limits_{\delta \rightarrow 0}\frac{\Eb_{\bmu}[\tau_\delta]}{\log(1/\delta)}\leq \theta T^*(\boldsymbol{\mu}).
\end{align*}
which is the result is \cite{garivier2016optimal}. Thus, our algorithm does improve from the previous ones and can handle more situations.

\subsection{Ranking Identification} 
Ranking identification is another important task. Instead of just identifying the best arm, the agent needs to identify the total order of all the $K$ arms in the bandits model $\bmu$. In our theory, considering the pure-exploration problem $(\mathcal{G}, \varphi)$ of identifying the rank of all the arms with respect to their expected rewards, $M = K!$ and each $\mathcal{G}_m \in \mathcal{G}$ is a singleton containing some $\sigma$. It is a pairwise exploration task since $\{\sigma\} = \cap_{i = 1}^{K-1} \mathcal{B}_{\sigma^{-1}(i)\sigma^{-1}(i+1)}$.
By taking different values of cost vector $\bc$, our theorem can be applied to minimize sample complexity setting and regret minimization setting, both of which are intriguing and fresh settings.

The auxiliary optimal value has an explicit expression in the ranking identification problem as illustrated in Lemma~\ref{lem:opt} below which can help us to calculate the value of $T^*(\boldsymbol{\Delta},\boldsymbol{\mu})$ in the theorem.
\begin{lemma}\label{lem:opt}
    Assume $\mu_1>\mu_2>\cdots>\mu_K$. For ranking identification task, $\supp(\calI_m)=\calA$, so $P_\calG(\bc,\bmu)=P(\bc)$ and $N_\calG(\bc,\bmu)=N(\bc)$. For every $\omega \in \Sigma_{P(\boldsymbol{c})}$, We have
    {
    \small
    \begin{align*}
        \inf\limits_{\boldsymbol{\lambda}\in \Alt(\boldsymbol{c},\boldsymbol{\mu})}(&\sum\limits_{a \in P(\boldsymbol{c})}\omega_a \frac{d(\mu_a,\lambda_a)}{c_a})=\min\bigg[\ \min\limits_{a> b,\ a,b\notin N(\boldsymbol{c})}\ (\frac{\omega_b}{c_b}+\frac{\omega_a}{c_a})I_{\frac{\omega_b/c_b}{\omega_b/c_b+\omega_a/c_a}}(\mu_b,\mu_a),\\
        &\min\limits_{a> b ,\ a\notin N(\boldsymbol{c}),\ b\in N(\boldsymbol{c})}\ \ (\frac{\omega_a}{c_a} d(\mu_a,\mu_b)),\ \min\limits_{a> b ,\ b\notin N(\boldsymbol{c}),\ a\in N(\boldsymbol{c})}\ (\frac{\omega_b}{c_b} d(\mu_b,\mu_a))\bigg].
    \end{align*}
    }
\end{lemma}
\begin{Proof}
    Let $\boldsymbol{\mu}$ such that $\mu_1>\mu_2>\ldots >\mu_K$ and we can note that $\Alt(\boldsymbol{c},\boldsymbol{\mu})=\mathop{\cup}\limits_{a>b}\{\boldsymbol{\lambda} \in S\ : \ \lambda_a>\lambda_b,\ \lambda_i=\mu_i\ \forall i\in N(\boldsymbol{c})\}$. Using that fact, one has
    \begin{align*}
        T^*(\boldsymbol{c},\boldsymbol{\mu})^{-1}&=\sup\limits_{\omega \in \Sigma_{P(\boldsymbol{c})}}\min\limits_{a>b}\inf\limits_{\boldsymbol{\lambda}\in S : \lambda_a>\lambda_b,\ \lambda_i=\mu_i\ \forall i\in N(\boldsymbol{c})}\bigg(\sum\limits_{a\in P(\boldsymbol{c})} \frac{\omega_a}{c_a} d(\mu_a,\lambda_a)\bigg)\\
        &=\sup\limits_{\omega \in \Sigma_{P(\boldsymbol{c})}}\min\limits_{a>b}\inf\limits_{\boldsymbol{\lambda}\in S : \lambda_a\geq\lambda_b,\ \lambda_i=\mu_i\ \forall i\in N(\boldsymbol{c})}\bigg(\sum\limits_{a\in P(\boldsymbol{c})} \frac{\omega_a}{c_a} d(\mu_a,\lambda_a)\bigg)\\
        &= \sup\limits_{\omega \in \Sigma_{P(\boldsymbol{c})}}\min\bigg[\ \min\limits_{a> b ,\ a,b\notin N(\boldsymbol{c})}\inf\limits_{\boldsymbol{\lambda}\in S : \lambda_a\geq\lambda_b}\bigg(\frac{\omega_b}{c_b} d(\mu_b,\lambda_b)+\frac{\omega_a}{c_a} d(\mu_a,\lambda_a)\bigg),\\
        &\quad\quad\quad\quad\ \ \ \ \ \ \ \ \min\limits_{a> b ,\ a\notin N(\boldsymbol{c}),\ b\in N(\boldsymbol{c})}\ \inf\limits_{\boldsymbol{\lambda}\in S : \lambda_a\geq\lambda_b=\mu_b}\bigg(\frac{\omega_a}{c_a} d(\mu_a,\lambda_a)\bigg),\\
        &\quad\quad\quad\quad\ \ \ \ \ \ \ \ \min\limits_{a> b ,\ b\notin N(\boldsymbol{c}),\ a\in N(\boldsymbol{c})}\ \inf\limits_{\boldsymbol{\lambda}\in S : \mu_a=\lambda_a\geq\lambda_b}\bigg(\frac{\omega_b}{c_b} d(\mu_b,\lambda_b)\bigg)\bigg]\\
        &= \sup\limits_{\omega \in \Sigma_{P(\boldsymbol{c})}}\min\bigg[\ \min\limits_{a> b ,\ a,b\notin N(\boldsymbol{c})}\ \inf\limits_{\boldsymbol{\lambda}\in S : \lambda_a\geq\lambda_b}\bigg(\frac{\omega_b}{c_b} d(\mu_b,\lambda_b)+\frac{\omega_a}{c_a} d(\mu_a,\lambda_a)\bigg),\\
        &\ \ \ \ \ \ \min\limits_{a> b ,\ a\notin N(\boldsymbol{c}),\ b\in N(\boldsymbol{c})} \bigg(\frac{\omega_a}{c_a} d(\mu_a,\mu_b)\bigg),\min\limits_{a> b ,\ b\notin N(\boldsymbol{c}),\ a\in N(\boldsymbol{c})}\ \bigg(\frac{\omega_b}{c_b} d(\mu_b,\mu_a)\bigg)\bigg]
    \end{align*}
    Minimizing
    \begin{align*}
        f(\lambda_b,\lambda_a)=\frac{\omega_b}{c_b}d(\mu_b,\lambda_b)+\frac{\omega_a}{c_a}d(\mu_a,\lambda_a)
    \end{align*}
    under the constraint $\lambda_a\geq \lambda_b$ is a convex optimization problem that can be solved analytically. The minimum is obtained for
    \begin{align*}
        \lambda_b=\lambda_a=\frac{\omega_b/c_b}{\omega_b/c_b+\omega_a/c_a}\mu_b+\frac{\omega_a/c_a}{\omega_b/c_b+\omega_a/c_a}\mu_a
    \end{align*}
    and its value can be written $(\frac{\omega_b}{c_b}+\frac{\omega_a}{c_a})I_{\frac{\omega_b/c_b}{\omega_b/c_b+\omega_a/c_a}}(\mu_b,\mu_a)$.\\
    So we have
    \begin{align*}
        T^*(\boldsymbol{c},\boldsymbol{\mu})^{-1}= \sup\limits_{\omega \in \Sigma_{P(\boldsymbol{c})}}\min\bigg[\ \min\limits_{a> b ,a,b\notin N(\boldsymbol{c})}\bigg(\frac{\omega_b}{c_b}+\frac{\omega_a}{c_a}\bigg)I_{\frac{\omega_b/c_b}{\omega_b/c_b+\omega_a/c_a}}(\mu_b,\mu_a),\\
        \min\limits_{a> b ,a\notin N(\boldsymbol{c}),\ b\in N(\boldsymbol{c})}\bigg(\frac{\omega_a}{c_a} d(\mu_a,\mu_b)\bigg),\min\limits_{a> b ,b\notin N(\boldsymbol{c}),\ a\in N(\boldsymbol{c})}\bigg(\frac{\omega_b}{c_b} d(\mu_b,\mu_a)\bigg)\bigg].
    \end{align*}
\end{Proof}

For some closed-form expressions, we give an example of ranking identification under regret minimization in three-armed bandit, we can present its closed-form clearly.
For a three-armed 1-Gaussian bandit model $\boldsymbol{\mu}=(\mu_1,\mu_2,\mu_3)$, $\boldsymbol{c}=\boldsymbol{\Delta}=(0,\Delta_2,\Delta_3)$, $T^*(\boldsymbol{\Delta},\boldsymbol{\mu})$ and $\boldsymbol{u}^*(\boldsymbol{\Delta},\boldsymbol{\mu})=G_{\boldsymbol{\Delta}}(\boldsymbol{\omega}^*(\boldsymbol{\Delta},\boldsymbol{\mu}))$ are shown in next proposition.
\begin{proposition}\label{pro:three-arm}
    Let $\delta\in (0,1)$. For any $\delta$-PAC algorithm under ranking identification with cost vector $\boldsymbol{\Delta}$. Any 3-armed 1-Gaussian bandits with reward expectations $\{\mu_1>\mu_2>\mu_3\}$ has:
    \begin{equation*}
        \boldsymbol{u}^*(\boldsymbol{\Delta},\boldsymbol{\mu})=\left\{
        \begin{array}{ccl}
             (0,\frac{\sqrt{\Delta_3}}{\sqrt{\Delta_2}+\sqrt{\Delta_3}},\frac{\sqrt{\Delta_2}}{\sqrt{\Delta_2}+\sqrt{\Delta_3}})& & {\Delta_3\leq [(3+\sqrt{5})/2]\Delta_2}  \\
             (0,\frac{\Delta_3^2-2\Delta_2\Delta_3}{(\Delta_3-\Delta_2)^2},\frac{\Delta_2^2}{(\Delta_3-\Delta_2)^2})& & {\Delta_3> [(3+\sqrt{5})/2]\Delta_2}
        \end{array}
        \right.
    \end{equation*}
    \begin{equation*}
        T^*(\boldsymbol{\Delta},\boldsymbol{\mu})^{-1}=\left\{
        \begin{array}{ccl}
             \frac{(\sqrt{\Delta_2}-\sqrt{\Delta_3})^2}{2}& & {\Delta_3\leq [(3+\sqrt{5})/2]\Delta_2}  \\
             \frac{\Delta_2}{2}\frac{\Delta_3-2\Delta_2}{\Delta_3-\Delta_2}& & {\Delta_3> [(3+\sqrt{5})/2]\Delta_2}
        \end{array}
        \right.
    \end{equation*}
    and $R(\tau_\delta)\geq T^*(\boldsymbol{\Delta},\boldsymbol{\mu})\kl(\delta,1-\delta)$.
\end{proposition}
\begin{Proof}
    First, consider the expression of $T^*(\boldsymbol{\Delta},\boldsymbol{\mu})^{-1}$ and the setting is $\mu_1>\mu_2>\mu_3,\ \Delta_i=\mu_1-\mu_i$ is the sub-optimal gap. Recall that $d(x,y)=(x-y)^2/2$ under 1-Gaussian bandit setting, so
    \begin{align*}
        T^*(\boldsymbol{\Delta},\boldsymbol{\mu})^{-1}&=\sup\limits_{\omega_2+\omega_3=1}\min\big[\frac{\omega_2}{\Delta_2}d(\mu_2,\mu_1),\frac{\omega_3}{\Delta_3}d(\mu_3,\mu_1),(\frac{\omega_2}{\Delta_2}+\frac{\omega_3}{\Delta_3})I_{\frac{\omega_2/\Delta_2}{\omega_2/\Delta_2+\omega_3/\Delta_3}}(\mu_2,\mu_3)\big]\\
        &=\sup\limits_{\omega_2+\omega_3=1}\min\big[\frac{\omega_2\Delta_2}{2},\frac{\omega_3\Delta_3}{2},(\frac{\omega_2}{\Delta_2}+\frac{\omega_3}{\Delta_3})I_{\frac{\omega_2\Delta_3}{\omega_2\Delta_3+\omega_3\Delta_2}}(\mu_2,\mu_3)\big].
    \end{align*}
    
    Noticing that     
    \begin{align*}
        (\frac{\omega_2}{\Delta_2}+\frac{\omega_3}{\Delta_3})I_{\frac{\omega_2/\Delta_2}{\omega_2/\Delta_2+\omega_3/\Delta_3}}&(\mu_2,\mu_3)=\inf\limits_{\lambda_3\geq \lambda_2}(\frac{\omega_2}{\Delta_2}d(\mu_2,\lambda_2)+\frac{\omega_3}{\Delta_3}d(\mu_3,\lambda_3))\\
        &\leq_{\lambda_3=\mu_1,\lambda_2=\mu_2}(\frac{\omega_2}{\Delta_2}d(\mu_2,\lambda_2)+\frac{\omega_3}{\Delta_3}d(\mu_3,\lambda_3))\\
        &=\frac{\omega_3}{\Delta_3}d(\mu_3,\mu_1),
    \end{align*}
 we do not need to take term $\frac{\omega_3}{\Delta_3}d(\mu_3,\mu_1)$ into consideration in computing $T^*(\boldsymbol{\Delta},\boldsymbol{\mu})^{-1}$. 
    Let $\omega_3=1-\omega_2$ and     
    \begin{align*}
        (\frac{\omega_2}{\Delta_2}+\frac{\omega_3}{\Delta_3})I_{\frac{\omega_2\Delta_3}{\omega_2\Delta_3+\omega_3\Delta_2}}(\mu_2,\mu_3)=\frac{(\mu_2-\mu_3)^2}{2}\frac{\omega_2(1-\omega_2)}{\Delta_3\omega_2+\Delta_2(1-\omega_2)}.
    \end{align*}
    Define function 
    \begin{align*}
        h(x)=\frac{(\mu_2-\mu_3)^2}{2}\frac{x(1-x)}{\Delta_3x+\Delta_2(1-x)},\ x\in [0,1].
    \end{align*}
    We need to discover some properties of $h(x)$
    \begin{align*}
        h'(x)=\frac{(\mu_2-\mu_3)^2}{2}\frac{-\Delta_3x^2+\Delta_2(x-1)^2}{(-\Delta_3x+\Delta_2(x-1))^2}
    \end{align*}
    and
    \begin{align*}
        h''(x)=\frac{(\mu_2-\mu_3)^2}{2}\frac{2\Delta_2\Delta_3}{(-\Delta_3x+\Delta_2(x-1))^3}.
    \end{align*}
    Since $-\Delta_3x+\Delta_2(x-1)<0$, so $h''(x)<0$ which means $h(x)$ is convex in $[0,1]$ and $h'(x)=0$ shows that $h(x)$ achieve maximum value $h_m=\frac{(\mu_2-\mu_3)^2}{2(\sqrt{\Delta_2}+\sqrt{\Delta_3})^2}=\frac{(\sqrt{\Delta_2}-\sqrt{\Delta_3})^2}{2}$ at point $x=\frac{\sqrt{\Delta_2}}{\sqrt{\Delta_2}+\sqrt{\Delta_3}}$.  After establishing those properties, we can continue our discussion.     When $h_m\leq \frac{\Delta_2}{2}\frac{\sqrt{\Delta_2}}{\sqrt{\Delta_2}+\sqrt{\Delta_3}}$ which means $(\mu_2-\mu_3)^2\leq \Delta_2^{3/2}(\sqrt{\Delta_2}+\sqrt{\Delta_3})$. Solving this inequality with condition $\Delta_3>\Delta_2$, we get $\Delta_3\leq [(3+\sqrt{5})/2]\Delta_2$, and further have:
    \begin{align*}
        T^*(\boldsymbol{\Delta},\boldsymbol{\mu})^{-1}=h_m=\frac{(\sqrt{\Delta_2}-\sqrt{\Delta_3})^2}{2}.
    \end{align*}
    When $h_m> \frac{\Delta_2}{2}\frac{\sqrt{\Delta_2}}{\sqrt{\Delta_2}+\sqrt{\Delta_3}}$, which means $\Delta_3> [(3+\sqrt{5})/2]\Delta_2$, let $x_0$ be the solution of $h(x)=\frac{\Delta_2}{2}x$ and we get:
    \begin{align*}
        T^*(\boldsymbol{\Delta},\boldsymbol{\mu})^{-1}=\frac{\Delta_2}{2}x_0
    \end{align*}
    where
    \begin{align*}
        x_0=\frac{(\mu_2-\mu_3)^2-\Delta_2^2}{(\mu_2-\mu_3)^2+\Delta_2\Delta_3-\Delta_2^2}=\frac{\Delta_3-2\Delta_2}{\Delta_3-\Delta_2}.
    \end{align*}
\end{Proof}

\subsection{Proof of Equation~\eqref{for:reg1} in the Main Paper}\label{appendix:proveequ}

Assume $\mu_1$ is the unique best arm, with confidence $\delta$ and $\varphi(\bmu)=1,\calI_1=\{(1,a):a=2,\ldots,K\}$, and $\supp(\calI(\bmu))=\calA$. Our theorem implies that
\begin{align*}
    R(\tau_\delta)\leq \theta T^*(\boldsymbol{\Delta},\bmu)\log\left(\frac{1}{\delta}\right)
\end{align*}
where $\boldsymbol{\Delta}= (0,\Delta_1,\ldots,\Delta_K)$ and using Lemma~\ref{lem:optgernal} in Appendix~\ref{sec:character}
\begin{align*}
    T^*(\boldsymbol{\Delta},\bmu)^{-1} = \sup\limits_{\omega_2+\cdots+\omega_K=1 \atop \omega_i > 0} \min\limits_{a\in\{2,\ldots,K\}}\left\{\frac{\omega_a}{\Delta_a} d(\mu_a,\mu_1)\right\}.
\end{align*}
To solve the final optimization problem, we use some inequalities: 
\begin{align*}
    \min\limits_{a\in\{2,\ldots,K\}}\left\{\frac{\omega_a}{\Delta_a} d(\mu_a,\mu_1)\right\}\leq& \sum\limits_{a=2}^K \frac{\omega_a}{\Delta_a} d(\mu_a,\mu_1)\cdot \frac{\frac{\Delta_a}{d(\mu_a,\mu_1)}}{\sum\limits_{i=2}^K \frac{\Delta_i}{d(\mu_i,\mu_1)}}\\
    =&\sum\limits_{a=2}^K\omega_a\cdot\frac{1}{\sum\limits_{i=2}^K \frac{\Delta_i}{d(\mu_i,\mu_1)}}.
\end{align*}
The first inequality comes from the fact that $\min\{a_1,\ldots,a_n\}\leq \sum_ip_ia_i$, if $\sum_ip_i=1,p_i\geq 0$. In the last term, we have $\sum_{a}\omega=1$, so
\begin{align*}
    \min\limits_{a\in\{2,\ldots,K\}}\left\{\frac{\omega_a}{\Delta_a} d(\mu_a,\mu_1)\right\}\leq \frac{1}{\sum\limits_{i=2}^K \frac{\Delta_i}{d(\mu_i,\mu_1)}}.
\end{align*}
The right-hand side is a constant and can be obtained by
\begin{align*}    (\omega_2,\ldots,\omega_K)=\left(\frac{\frac{\Delta_2}{d(\mu_2,\mu_1)}}{\sum\limits_{i=2}^K \frac{\Delta_i}{d(\mu_i,\mu_1)}},\ldots,\frac{\frac{\Delta_K}{d(\mu_K,\mu_1)}}{\sum\limits_{i=2}^K \frac{\Delta_i}{d(\mu_i,\mu_1)}}\right).
\end{align*}
So, we have 
\begin{align*}
    T^*(\boldsymbol{\Delta},\bmu)^{-1} = \sup\limits_{\omega_2+\cdots+\omega_K=1 \atop \omega_i > 0} \min\limits_{a\in\{2,\ldots,K\}}\left\{\frac{\omega_a}{\Delta_a} d(\mu_a,\mu_1)\right\}=\frac{1}{\sum\limits_{i=2}^K \frac{\Delta_i}{d(\mu_i,\mu_1)}},
\end{align*}
and the regret in our ETC algorithm is 
\begin{align*}
    R_{\bmu}(T)\leq \underbrace{\theta T^*(\boldsymbol{\Delta},\bmu)\log(T)}_{\text{regret before commitment}}+\underbrace{(T-O(\log^{1+r}(T)))\frac{1}{T}\max_a{\Delta_a}}_{\text{regret during commitment}}.
\end{align*}
The last term can be bounded by a constant, and we get
\begin{align*}
    R_{\bmu}(T)\leq \theta\sum\limits_{i=2}^K \frac{\Delta_i}{d(\mu_i,\mu_1)}\log(T) = \theta\sum\limits_{i=2}^K \frac{\Delta_i}{\KL(\mu_i,\mu_1)}\log(T)
\end{align*}
where $\theta$ can  be chosen to be close to $1$, which concludes the proof.

\end{document}

%% file: preamble.tex
\usepackage{amssymb,amsmath,amsfonts,bm,epsfig,graphicx,theorem}
\usepackage{xspace}
\usepackage{cleveref}

\newcommand{\comment}[1]{}

\usepackage{xspace}

\newcommand{\Rb}{\mathbb{R}}

\newcommand{\Eb}{\mathbb{E}}
\newcommand{\Pb}{\mathbb{P}}

\newcommand{\Nb}{\mathbb{N}}

\newcommand{\lv}{\mathbf{1}}

\newcommand{\bmu}{\boldsymbol{\mu}}
\newcommand{\blambda}{\boldsymbol{\lambda}}

\newcommand{\bc}{\boldsymbol{c}}
\newcommand{\bu}{\boldsymbol{u}}

\newcommand{\calA}{\mathcal{A}}
\newcommand{\calB}{\mathcal{B}}

\newcommand{\calE}{\mathcal{E}}
\newcommand{\calF}{\mathcal{F}}
\newcommand{\calG}{\mathcal{G}}
\newcommand{\calH}{\mathcal{H}}
\newcommand{\calI}{\mathcal{I}}

\newcommand{\calR}{\mathcal{R}}

\newcommand{\argmax}{\mathop{\mbox{\rm arg\,max}}}

\newtheorem{theorem}{Theorem}

\newtheorem{lemma}{Lemma}
\newtheorem{proposition}{Proposition}
\newtheorem{definition}{Definition}

\newtheorem{remark}{Remark}
\newtheorem{example}{Example}
\newenvironment{Proof}[1]{\medskip\par\noindent
{\bf Proof:\,}\,#1}{{\mbox{\,$\blacksquare$}\par}}
\newenvironment{Proofsketch}[1]{\medskip\par\noindent
{\bf Proof sketch:\,}\,#1}{{\mbox{\,$\blacksquare$}\par}}

%% file: revised_aistats_paper.bbl
\begin{thebibliography}{}

\bibitem[Audibert and Bubeck, 2010]{audibert2010best}
Audibert, J.-Y. and Bubeck, S. (2010).
\newblock Best arm identification in multi-armed bandits.
\newblock In {\em Conference on Learning Theory}, pages 13--p.

\bibitem[Auer et~al., 2002]{auer2002finite}
Auer, P., Cesa-Bianchi, N., and Fischer, P. (2002).
\newblock Finite-time analysis of the multiarmed bandit problem.
\newblock {\em Machine learning}, 47:235--256.

\bibitem[Aziz et~al., 2021]{aziz2021multi}
Aziz, M., Kaufmann, E., and Riviere, M.-K. (2021).
\newblock On multi-armed bandit designs for dose-finding trials.
\newblock {\em Journal of Machine Learning Research}, 22(14):1--38.

\bibitem[Cappé et~al., 2013]{Capp__2013}
Cappé, O., Garivier, A., Maillard, O.-A., Munos, R., and Stoltz, G. (2013).
\newblock Kullback–leibler upper confidence bounds for optimal sequential allocation.
\newblock {\em The Annals of Statistics}, 41(3).

\bibitem[Carpentier and Locatelli, 2016]{carpentier2016tight}
Carpentier, A. and Locatelli, A. (2016).
\newblock Tight (lower) bounds for the fixed budget best arm identification bandit problem.
\newblock In {\em Conference on Learning Theory}, pages 590--604. PMLR.

\bibitem[Chen et~al., 2017a]{chen2017nearlyoptimalsamplingalgorithms}
Chen, L., Gupta, A., Li, J., Qiao, M., and Wang, R. (2017a).
\newblock Nearly optimal sampling algorithms for combinatorial pure exploration.

\bibitem[Chen et~al., 2017b]{chen2017nearly}
Chen, L., Li, J., and Qiao, M. (2017b).
\newblock Nearly instance optimal sample complexity bounds for top-k arm selection.
\newblock In {\em Artificial Intelligence and Statistics}, pages 101--110. PMLR.

\bibitem[Du et~al., 2022]{du2022duelingbanditstwoduelingmultidueling}
Du, Y., Wang, S., and Huang, L. (2022).
\newblock Dueling bandits: From two-dueling to multi-dueling.

\bibitem[Dud{\'\i}k et~al., 2015]{dudik2015contextual}
Dud{\'\i}k, M., Hofmann, K., Schapire, R.~E., Slivkins, A., and Zoghi, M. (2015).
\newblock Contextual dueling bandits.
\newblock In {\em Conference on Learning Theory}, pages 563--587. PMLR.

\bibitem[Even-Dar et~al., 2006]{even2006action}
Even-Dar, E., Mannor, S., Mansour, Y., and Mahadevan, S. (2006).
\newblock Action elimination and stopping conditions for the multi-armed bandit and reinforcement learning problems.
\newblock {\em Journal of Machine Learning Research}, 7(6).

\bibitem[Gabillon et~al., 2012]{gabillon2012best}
Gabillon, V., Ghavamzadeh, M., and Lazaric, A. (2012).
\newblock Best arm identification: A unified approach to fixed budget and fixed confidence.
\newblock {\em Advances in Neural Information Processing Systems}, 25.

\bibitem[Gabillon et~al., 2011]{gabillon2011multi}
Gabillon, V., Ghavamzadeh, M., Lazaric, A., and Bubeck, S. (2011).
\newblock Multi-bandit best arm identification.
\newblock {\em Advances in Neural Information Processing Systems}, 24.

\bibitem[Gai et~al., 2010]{gai2010learning}
Gai, Y., Krishnamachari, B., and Jain, R. (2010).
\newblock Learning multiuser channel allocations in cognitive radio networks: A combinatorial multi-armed bandit formulation.
\newblock In {\em 2010 IEEE Symposium on New Frontiers in Dynamic Spectrum (DySPAN)}, pages 1--9. IEEE.

\bibitem[Garivier and Kaufmann, 2016]{garivier2016optimal}
Garivier, A. and Kaufmann, E. (2016).
\newblock Optimal best arm identification with fixed confidence.
\newblock In {\em Conference on Learning Theory}, pages 998--1027. PMLR.

\bibitem[Jedra and Proutiere, 2020]{jedra2020optimal}
Jedra, Y. and Proutiere, A. (2020).
\newblock Optimal best-arm identification in linear bandits.
\newblock {\em Advances in Neural Information Processing Systems}, 33:10007--10017.

\bibitem[Kalyanakrishnan and Stone, 2010]{kalyanakrishnan2010efficient}
Kalyanakrishnan, S. and Stone, P. (2010).
\newblock Efficient selection of multiple bandit arms: Theory and practice.
\newblock In {\em ICML}, volume~10, pages 511--518.

\bibitem[Kalyanakrishnan et~al., 2012]{kalyanakrishnan2012pac}
Kalyanakrishnan, S., Tewari, A., Auer, P., and Stone, P. (2012).
\newblock Pac subset selection in stochastic multi-armed bandits.
\newblock In {\em ICML}, volume~12, pages 655--662.

\bibitem[Kanarios et~al., 2024]{kanarios2024cost}
Kanarios, K., Zhang, Q., and Ying, L. (2024).
\newblock Cost aware best arm identification.
\newblock {\em arXiv preprint arXiv:2402.16710}.

\bibitem[Kaufmann et~al., 2016]{kaufmann2016complexity}
Kaufmann, E., Capp{\'e}, O., and Garivier, A. (2016).
\newblock On the complexity of best-arm identification in multi-armed bandit models.
\newblock {\em The Journal of Machine Learning Research}, 17(1):1--42.

\bibitem[Komiyama et~al., 2022]{komiyama2022minimax}
Komiyama, J., Tsuchiya, T., and Honda, J. (2022).
\newblock Minimax optimal algorithms for fixed-budget best arm identification.
\newblock {\em Advances in Neural Information Processing Systems}, 35:10393--10404.

\bibitem[Lattimore and Szepesv{\'a}ri, 2020]{lattimore2020bandit}
Lattimore, T. and Szepesv{\'a}ri, C. (2020).
\newblock {\em Bandit algorithms}.
\newblock Cambridge University Press.

\bibitem[Li et~al., 2010]{li2010contextual}
Li, L., Chu, W., Langford, J., and Schapire, R.~E. (2010).
\newblock A contextual-bandit approach to personalized news article recommendation.
\newblock In {\em Proceedings of the 19th International Conference on World Wide Web}, pages 661--670.

\bibitem[Liu et~al., 2021]{liu2021pond}
Liu, X., Li, B., Shi, P., and Ying, L. (2021).
\newblock Pond: Pessimistic-optimistic online dispatching.

\bibitem[Magureanu et~al., 2014]{Magureanu2014}
Magureanu, S., Combes, R., and Prouti{\'e}re, A. (2014).
\newblock Lipschitz bandits: Regret lower bounds and optimal algorithms.
\newblock {\em Proceedings of the 27th Conference on Learning Theory}.

\bibitem[Qin et~al., 2025]{qin2025cost}
Qin, Z., Xue, W., Zheng, L., Gan, X., Wu, H., Jin, H., and Fu, L. (2025).
\newblock Cost-aware best arm identification in stochastic bandits.
\newblock {\em ACM Transactions on Intelligent Systems and Technology}.

\bibitem[Shen, 2019]{shen2019universal}
Shen, C. (2019).
\newblock Universal best arm identification.
\newblock {\em IEEE Transactions on Signal Processing}, 67(17):4464--4478.

\bibitem[Shin et~al., 2019]{shin2019bias}
Shin, J., Ramdas, A., and Rinaldo, A. (2019).
\newblock On the bias, risk and consistency of sample means in multi-armed bandits.
\newblock {\em arXiv preprint arXiv:1902.00746}.

\bibitem[Soare et~al., 2014]{soare2014best}
Soare, M., Lazaric, A., and Munos, R. (2014).
\newblock Best-arm identification in linear bandits.
\newblock {\em Advances in Neural Information Processing Systems}, 27.

\bibitem[Thompson, 1933]{thompson1933likelihood}
Thompson, W.~R. (1933).
\newblock On the likelihood that one unknown probability exceeds another in view of the evidence of two samples.
\newblock {\em Biometrika}, 25(3-4):285--294.

\bibitem[Wang et~al., 2024]{wang2024best}
Wang, P.-A., Tzeng, R.-C., and Proutiere, A. (2024).
\newblock Best arm identification with fixed budget: A large deviation perspective.
\newblock {\em Advances in Neural Information Processing Systems}, 36.

\bibitem[Yue et~al., 2012]{yue2012k}
Yue, Y., Broder, J., Kleinberg, R., and Joachims, T. (2012).
\newblock The k-armed dueling bandits problem.
\newblock {\em Journal of Computer and System Sciences}, 78(5):1538--1556.

\bibitem[Zhang and Ying, 2024]{zhang2023fast}
Zhang, Q. and Ying, L. (2024).
\newblock Fast and regret optimal best arm identification: fundamental limits and low-complexity algorithms.
\newblock {\em Advances in Neural Information Processing Systems}, 36.

\bibitem[Zhou and Tian, 2022]{zhou2022approximate}
Zhou, R. and Tian, C. (2022).
\newblock Approximate top-$ m $ arm identification with heterogeneous reward variances.
\newblock In {\em International Conference on Artificial Intelligence and Statistics}, pages 7483--7504. PMLR.

\end{thebibliography}
